\documentclass[10pt,journal,compsoc]{IEEEtran} 
\usepackage[compress]{cite}
\usepackage{amsmath,amsfonts}
\usepackage{algorithmic}
\usepackage{algorithm}
\usepackage{array}
\usepackage{textcomp}
\usepackage{stfloats}
\usepackage{url}
\usepackage{verbatim}
\usepackage{graphicx}
\usepackage{cite}
\usepackage{amsmath}
\usepackage{amssymb}
\usepackage{booktabs}
\usepackage{multirow}
\usepackage{diagbox}
\usepackage{bbm}
\usepackage{hyperref} 
\usepackage{caption}
\usepackage{subcaption}
\usepackage{tabularx}
\usepackage[table]{xcolor}
\usepackage{float}
\usepackage{hyperref}

\begin{document}
\title{Predicting Genetic Mutation from Whole Slide Images via Biomedical-Linguistic Knowledge Enhanced Multi-label Classification}

\author{Gexin Huang, Chenfei Wu, Mingjie Li, Xiaojun Chang~\IEEEmembership{Senior~Member,~IEEE}, Ling Chen~\IEEEmembership{Senior~Member,~IEEE}, Ying Sun, Shen Zhao\IEEEauthorrefmark{1}, Xiaodan Liang, and Liang Lin~\IEEEmembership{Fellow,~IEEE}.

\thanks{G. Huang is with the University of British Columbia, Vancouver, Canada. C. Wu and Y. Sun are with Sun Yat-sen University Cancer Center, Guangzhou, 510060, China. M. Li is with the Radiation Oncology, at Stanford University. L. Chen and X. Chang are with the Australian Artificial Intelligence Institute, University of Technology Sydney. S. Zhao and X. Liang are with the School of Intelligent Systems Engineering, Sun Yat-sen University, Guangzhou, China, 510006. L. Lin is with the School of Computer Science and Engineering, Sun Yat-sen University, Guangzhou, China, 510006. X. Chang and X. Liang are also with the Computer Vision Department, Mohamed bin Zayed University of Artificial Intelligence (MBZUAI). (Email: gexinml@gmail.com, wucf@sysucc.org.cn, lmj695@stanford.edu, xiaojun.chang@uts.edu.au, Ling.Chen@uts.edu.au, sunying@sysucc.org.cn, z-s-06@163.com, xdliang328@gmail.com, linliang@ieee.org)}
\thanks{Corresponding author: Shen Zhao.}
\thanks{This work is supported by the National Natural Science Foundation of China under Grants 62101607, the National Key Research and Development Program Inter-governmental Special Project for International Science and Technology Innovation Cooperation under grants 2022YFE0112500, and the China Postdoctoral Science Foundation under Grants 2022TQ0389.}% <-this % stops a space
% \thanks{M. Li, X. Chang and L. Chen is with University of Technology Sydney. X. Liang is with Sun Yat-sen University, Shenzhen, 510521, China. C. Fei and Y. Sun is with Sun Yat-sen University Cancer Center, Guangzhou, 510060, China (* Corresponding author: Xiaodan Liang. E-mail: XXX, Shen Zhao. E-mail: XXX)}	
}
\markboth{IEEE Transactions on Pattern Analysis and Machine Intelligence}%
{Shell \MakeLowercase{\textit{et al.}}: A Sample Article Using IEEEtran.cls for IEEE Journals}

\IEEEpubid{0000--0000/00\$00.00~\copyright~2021 IEEE}

\IEEEtitleabstractindextext{
\begin{abstract}

Predicting genetic mutations from whole slide images is indispensable for cancer diagnosis. However, existing work training multiple binary classification models faces two challenges: (a) Training multiple binary classifiers is inefficient and would inevitably lead to a class imbalance problem. (b) The biological relationships among genes are overlooked, which limits the prediction performance. To tackle these challenges, we innovatively design a Biological-knowledge enhanced PathGenomic multi-label Transformer to improve genetic mutation prediction performances. 
BPGT first establishes a novel gene encoder that constructs gene priors by two carefully designed modules: (a) A gene graph whose node features are the genes' linguistic descriptions and the cancer phenotype, with edges modeled by genes' pathway associations and mutation consistencies. (b) A knowledge association module that fuses linguistic and biomedical knowledge into gene priors by transformer-based graph representation learning, capturing the intrinsic relationships between different genes' mutations.
BPGT then designs a label decoder finally that performs genetic mutation prediction by two tailored modules: (a) A modality fusion module that firstly fuses the gene priors with critical regions in WSIs and obtains gene-wise mutation logits. (b) A comparative multi-label loss that emphasizes the inherent comparisons among mutation status to enhance the discrimination capabilities.
Sufficient experiments on The Cancer Genome Atlas benchmark demonstrate that BPGT outperforms the state-of-the-art.

\end{abstract}

}

\maketitle

\begin{figure*}[htbp]
	\centering
	\includegraphics[width=\linewidth]{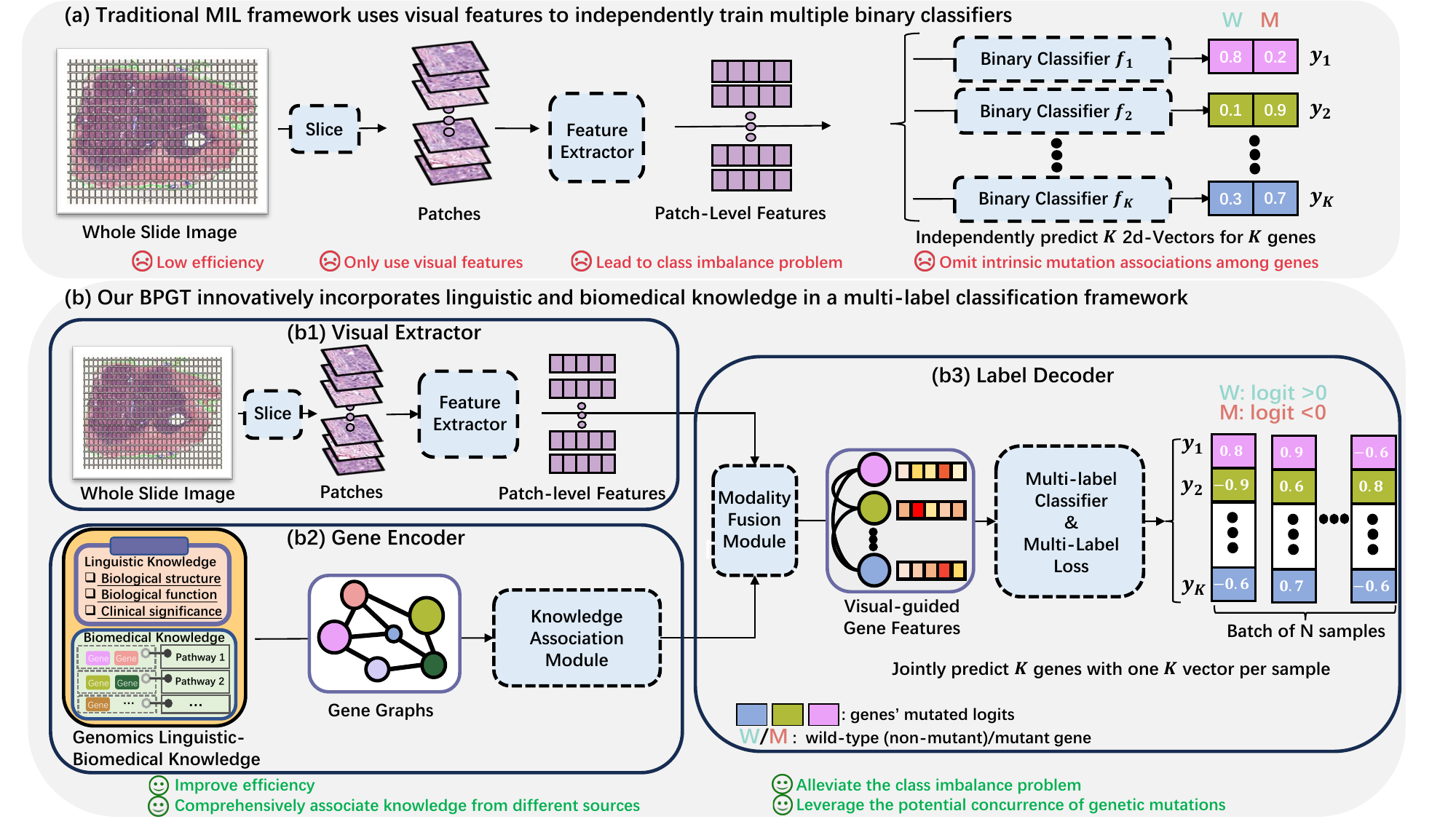}
         \caption{Comparison of the flowcharts of our BPGT with existing MIL frameworks. While MIL frameworks (Fig. \ref{fig_1} (a)) use visual features to independently predict 2D vectors for each gene indicating its mutation, our BPGT (Fig. \ref{fig_1} (b)) comprehensively associates knowledge from different sources (i.e., linguistic and biomedical knowledge) in a multi-label classification paradigm, which improves the efficiency, alleviates the class imbalance problem, fully leverages the potential concurrence of genetic mutations, and improves the feature discriminability.}
	\label{fig_1}
	\vspace{-0.3cm}
\end{figure*}

\section{Introduction}
\label{sec:intro}
% 思路： 要证明 多标签分类是重要的-》 
% 我们在做什么事情，目前有什么样的解法
% 这个解法目前有什么缺陷 
% 我们提出了什么样的方法
Predicting genetic mutations from whole slide images (WSIs), i.e., finding all mutated genes from the histopathology information in the input WSIs, holds significant promise in advancing clinical procedures for cancer diagnosis~\cite{lu2021ai}, prognosis~\cite{fu2020pan, ning2023survial}, survival prediction~\cite{di2023survialpred} and treatment~\cite{kather2019deep}. Predicting genetic mutations is clinically important because mutations in certain genes are inherently linked with the progression of cancer~\cite{martinez2020compendium}. Recently, researchers have unveiled the association between genetic mutations and histopathology information of cancer tissues~\cite{shia2017morphological}, which supports the feasibility of predicting genetic mutations directly from WSIs. Although genetic testing can serve as an alternative method, predicting genetic mutations directly from WSIs is significantly more cost-effective and convenient~\cite{mateo2022delivering}. Therefore, this approach represents a crucial direction for both clinical endeavors and artificial intelligence research. % 加个参考文献，论证一下直接做基因测试比较贵？√ 前面应用的文献中就表达这个意思了

However, predicting genetic mutations from WSIs is challenging: on one hand, genetic mutations represent alterations at the molecular level; their relationships with the visual histopathology information in WSIs are not easily discernible. On the other hand, the biological relationships of different gene mutations are also complex. For example: (1) The WSIs are typical of large sizes (often at the gigapixel level), however, the "fine" hints of which parts in them are related to the mutation of specific genes be impractical for even professional pathologists to discern, as the cellular and histological structures linked to genetic mutations are nuanced and intricate\cite{kather2020pan}. Thus, it could be even more challenging for computational models to explore the potential relationships between critical parts of gigapixel-level WSI and specific gene mutations. (2) To make it more difficult, the mutations of different genes could be subtly related at the molecular level \cite{chan2015clusters}. In other words, it is common for an individual patient to involve mutations in not just one gene but rather for concurrence mutations in various genes\cite{chalmers2017analysis}. These molecular-level biological relationships and concurrences in gene mutations are difficult for computer vision algorithms to capture. Thus, predicting genetic mutation from WSIs is in essence a multi-label classification task with complicated inputs and easily-confused class labels. 

Existing research mainly decomposes genetic mutation prediction into predicting binary mutation status for each gene (Fig. \ref{fig_1} (a)), which may be inefficient and inaccurate since they ignore the intrinsic biological relationships between the mutation of different genes \cite{schneider2022integration, karlebach2008modelling}. Furthermore, they may overlook the guidance from linguistic gene labels and biomedical knowledge, impeding the efficacy of deep learning in predicting genetic mutations from WSIs \cite{mohamed2021biological}. Detailed analyses are as follows: 

% 关键性--难点--现有工作尚无法解决这个问题
% 按照“相关工作先行”的思路，把思路整理完毕。后续也可以把介绍MIL的那一句放到下面1.1.1节。
% 以及，genetic alternations和mutations是不是同一个意思？→→统一写成mutations。

\subsection{Related work}
\subsubsection{Multi-instance learning (MIL) paradigm for genetic mutation prediction in WSIs.}
The MIL-based methods can be potentially used for predicting gene mutation, however, directly adopting this gene-wise binary mutation prediction paradigm may ignore the intrinsic gene mutation relationships and the information from the non-visual modalities. The MIL paradigm first divides the WSIs into multiple patches, then selects some representative patches and extracts their patch-level features, and lastly aggregates the features as slide-level features to complete the pathology diagnosis tasks. The MIL paradigm has proven its effectiveness in handling WSIs for different pathology diagnosis tasks (such as cancer diagnosis, molecular phenotype prediction, and survival prediction) \cite{guo2022robust, chen2023optimization, shengrui2023ssl}. The MIL paradigm has also been applied in predicting genetic mutation. For example, 
Fu \textit{et al.}\cite{fu2020pan} adopts a transfer-learning-based MIL method to classify the genetic mutation. They first use a pre-trained InceptionNet to extract transferable patch-level features, then sequentially aggregate the features to train a new classifier for the genetic mutation. 
Kather \textit{et al.}\cite{kather2020pan} first extracts patch-level features via a ResNet-50 encoder, then aggregates those features using a three-layer multi-layered perception (MLP) and an average pooling layer, and finally trains an independent binary-classifier for each gene. 
Qu \textit{et al.} \cite{qu2021genetic} first adopts a ResNet-101 encoder to extract patch-level features, then selects patches containing tumors using the K-means algorithm with the help of a human expert. They subsequently leverage self-attention layers for feature aggregation to learn the slide-level representation, which is lastly used to predict the genetic mutation. 
Chen \textit{et al.} \cite{chen2023optimization,guo2022robust} divides patch-level features into multiple clusters and trains corresponding classifiers respectively, accordingly calculating the slide-level prediction by choosing the best score among the classifiers. 
Saldanha \textit{et al.} \cite{saldanha2023self} first extracts patch-level features, then aggregates patch-level features via attention layers to score the genetic mutation.
Although these works achieve non-negligible achievements, they decompose the genetic mutation task into individually training binary classifiers for each gene to predict whether it is mutated (each classifier is responsible for predicting a two-dimensional vector for indicating the mutation of one gene, as shown in Fig. \ref{fig_1} (a), which could be further improved in accuracy and efficiency by better exploiting the genetic mutation relationships and multi-modality knowledge in a multi-label classification manner.
% 时态问题，全部统一现在时。
% mutations还是mutation？？---统一用mutation！

\subsubsection{Knowledge graph for medical images}
Leveraging knowledge graphs is promising to prompt genetic mutation prediction performance in WSI because they could potentially model the relationships between the mutation of different genes by semantically connecting knowledge using nodes and edges\cite{2023knowledgegraph}. Recent knowledge graph-based methods mainly view the images/WSI patches (containing tissues/lesions of interest) as nodes and the relationships among them (such as spatial distance and feature similarity) as edges, which focus on encoding better node feature representations and capturing better interactions among them.
For example, Chan \cite{chan2023histopathology} \textit{et al.} first detects nuclei in the WSI patches and classifies them into predefined nuclei types (e.g., neoplastic or non-neoplastic). Then, for each nuclei type, a graph is constructed using patch-level features as nodes and feature similarity as edges. A semantic-consistent pooling is used to jointly aggregate graphs and yield the WSI-level features for downstream tasks such as cancer classification. 
Li \cite{li2023semiwsi} \textit{et al.} first extracts disease-relevant features from WSI patches as nodes. Then, hierarchical graphs are constructed using the k-nearest neighbor (KNN) according to their spatial distances. Finally, the weighted pooling is used to aggregate node predictions from hierarchical graphs for cancer classification.
Mao~\cite{mao2022imagegcn} \textit{et al.} adopts a GCN framework that defines each X-ray image as a node and encodes four types of image relations (identity, age, gender, and view relations) as edges. The global image representations obtained by the GCN are then used for identifying 14 different diseases.
Yu \cite{yu2021cgnet} \textit{et al.} first extracts features from chest X-ray images as nodes, then encodes Euclidean distance between them as edges. Next, the nodes are aggregated through a GCN to build the global representations for all X-ray images, which are leveraged to classify each X-ray image into normal and pneumonia.
Although these methods may be transferred to genetic mutation prediction to improve the MIL methods by considering the mutation relationships, these knowledge graphs are mainly based on visual features and do not make full use of the valuable label information (e.g., the linguistic and biomedical information behind the labels), which may limit their ability in genetic mutation prediction. 

\subsubsection{Discussion and motivation}
Existing genetic mutation prediction methods could be further improved by better exploiting multi-modality information in a multi-label classification framework. Simply adopting the existing MIL diagram and/or the knowledge graphs in genetic mutation prediction could involve the following issues:
\vspace{-0.15cm}
\begin{itemize}
\item Since there could be large numbers of genes to be classified, individually training binary classifiers for each gene needs to train a large number of classifiers in a gene-wise manner. This results in a large number of parameters for training models and lowers the efficiency of the MIL framework \cite{cifci2022artificial}. More importantly, there are many more negative samples (not mutated genes) than positive ones (mutated ones) when training binary classifiers per gene; the trained classifiers would thus tend to yield negative results, which would be undesirable for further diagnosis and analysis.

\item Another drawback of decomposing genetic mutation prediction to individually training binary classifiers is that this strategy ignores the valuable intrinsic relationships in genomics data. However, in the realm of human genomics, the genes could be subtly related to each other at the molecular level, which means there may be potential concurrence of genetic mutations. Ignoring this fact may lead to sub-optimal genetic mutation prediction performance, especially for the rare genes with strong mutation relationships (i.e., the genes with insufficient training data, yet their mutation status can be inferred from related genes) \cite{schneider2022integration, karlebach2008modelling}.
\item Although knowledge graphs might be able to perform multi-label classification tasks by regarding the genes (rather than the WSI patches as in \cite{zhao2020predicting, ding2020feature}) as nodes, existing knowledge graph-based methods~\cite{li2022video} may ignore the valuable linguistic knowledge and biomedical knowledge of the genes. Thus, directly transferring knowledge graph-based methods to genetic mutation prediction may lack guidance from these comprehensive gene priors, which limits the performance of accurately locating the regions associated with genetic mutations in gigapixel-sized WSI images, consequently harming the mutation prediction performance. 
\end{itemize}
\vspace{-0.15cm}
Thus, a multi-label classification framework that comprehensively leverages the multi-modality information from the WSI images, the linguistic gene label knowledge, and the biomedical relationships between different genes is required for genetic mutation prediction. Furthermore, approaches should be considered to improve the feature discrimination for correctly finding all mutated genes.

\subsection{Overview of the proposed method}
Based on the above discussions, we propose a Biological-knowledge enhanced PathGenomic multi-label Transformer (BPGT) for predicting genetic mutations from WSIs in a multi-label classification manner. As shown in Fig. \ref{fig_1} (b), BPGT first extracts visual features as in previous work (Fig. \ref{fig_1} (b1)); what is different, it carefully designs a novel gene encoder (GE, Fig. \ref{fig_1} (b2)) based on GNN and transformers to integrate biomedical knowledge with linguistic knowledge as gene priors; it also designs a new label decoder (LD, Fig. \ref{fig_1} (b3)) that comprehensively fuses visual features and gene priors; it further formulates a comparative multi-label loss for multi-label classification: 

(1) The gene encoder is designed to integrate biomedical knowledge with linguistic knowledge to enhance genetic mutation prediction. As shown in Fig. \ref{fig_1} (b2), the gene encoder first designs a gene graph (GG) that combines linguistic knowledge from the GeneCard (such as the text description of the gene characteristics concerning gene mutation and cancer morbidity) and biomedical knowledge (such as cancer phenotypes, pathway associations, and gene mutation consistencies that reflect the relationships of different gene mutations and/or cancer morbidity), which forms a comprehensive description of the genes. Then, the gene encoder designs a knowledge association module (KAM) to fuse the linguistic knowledge and the biomedical knowledge into gene priors by transformer-based graph representation learning for capturing the intrinsic relationships between the mutation of different genes.  

(2) The label decoder is designed for fusing the gene priors with the visual features to link the linguistic knowledge and biomedical relationships with the WSI features. As shown in Fig. \ref{fig_1} (b3), the label decoder first designs a transformer-based modality fusion module (MFM) for multi-modality information fusion, which helps BPGT to focus on the visual features of the WSI parts that are the most relevant to the mutated genes. The label decoder then trains a multi-label classifier for each WSI (i.e., it predicts a \textit{N}-dimensional vector for indicating the mutation probability for the \textit{N} genes, as shown in the rightmost part of Fig. \ref{fig_1} (b3)).
Furthermore, the label decoder designs a comparative multi-label loss to better discriminate mutated genes from non-mutated ones via emphasizing inherent comparisons between classes, which avoids the need to train separate gene classifiers and enhances genetic mutation prediction performance. 

The major contributions of this paper are : 

\vspace{-0.15cm}
\begin{itemize}
% multi-label to incorporate the label relationships
\item To the best of our knowledge, our BPGT is the first multi-label classification framework for genetic mutation, which comprehensively integrates WSI visual features, linguistic knowledge, and biomedical knowledge of different genes. This design avoids the limitations of existing MIL-based genetic mutation work.

\item A gene encoder is proposed to integrate the linguistic label knowledge with the biomedical knowledge in a transformer-based graph representation learning manner, which constructs gene priors
to comprehensively describe the genes' biological characteristics.

\item A label decoder is designed to integrate the visual features and gene priors, which comprehensively leverages multi-modality knowledge to explore the associations between gigapixel-level WSI features and gene priors. It also carefully designs a new comparative multi-label loss to improve the genetic mutation prediction performance for multi-label classification.
\end{itemize}
\vspace{-0.15cm}

Comprehensive experiments on a challenging large-scale WSI dataset (The Cancer Genome Atlas, TCGA) are carried out for evaluation. It is demonstrated that our BPGT improves the performance in predicting genetic mutations and outperforms the state-of-the-art (SOTA) binary classification models. All designed modules are evaluated to be beneficial for genetic mutation prediction. Our BPGT lays the foundation for the research of mutation-related cancer onsets, targeted therapy, and prognoses.

\begin{figure*}[htbp]
	\centering
	\includegraphics[width=0.95\textwidth]{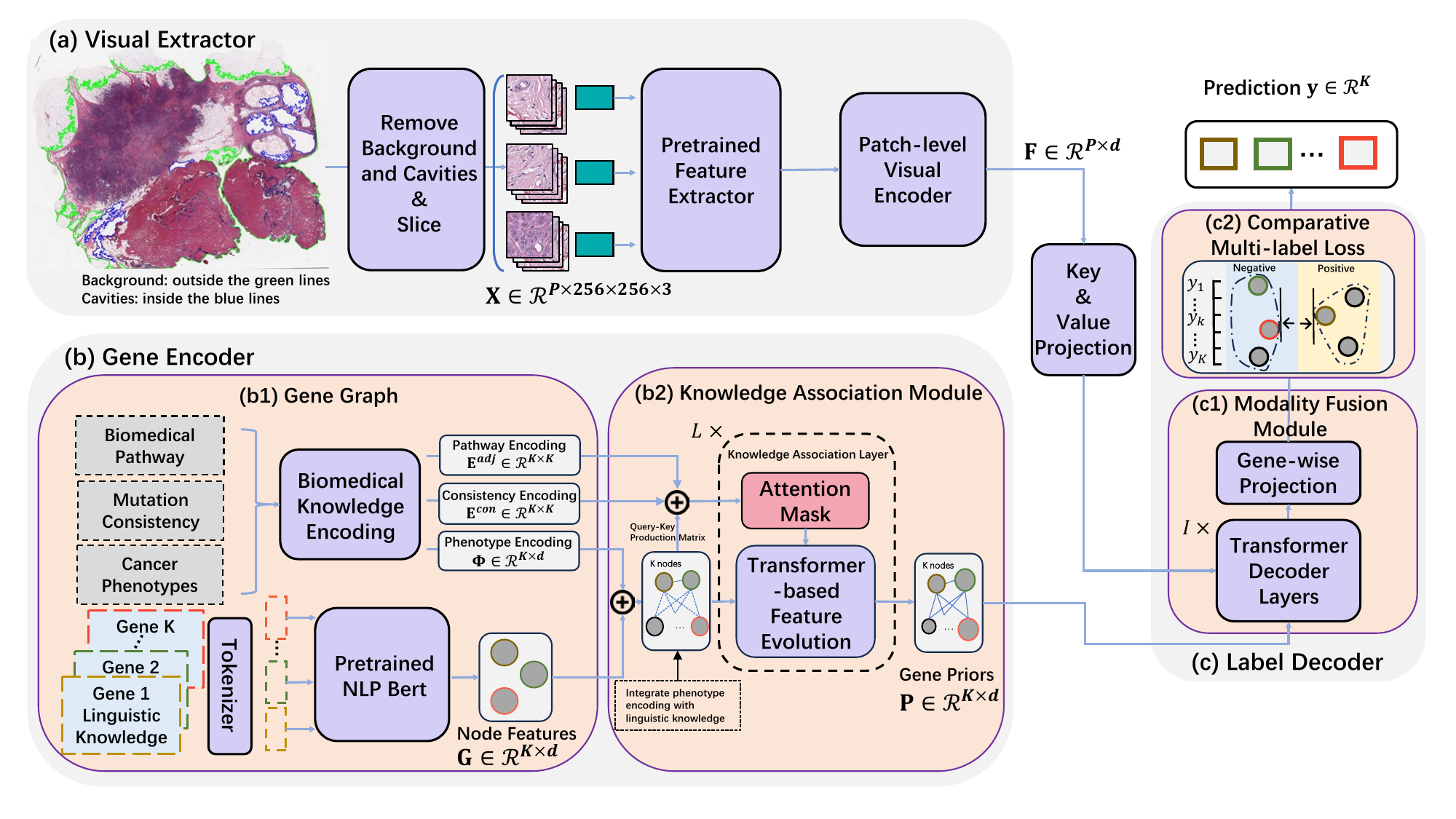}
	\vspace{-0.4cm}
	\caption{Illustration of the overall architecture of BPGT, which includes: a) visual extractor; b) gene encoder; c) label decoder. Details of (b1) gene graph (GG) and (b2) Knowledge Association Module (KAM) are respectively illustrated in \ref{mtd:main:geg} and \ref{mtd:main:kal}.}
	% 图的caption得改。230912
	\label{fig_2}
    \vspace{-0.5cm}
\end{figure*}

\section{BPGT}

BPGT is a tightly integrated multi-label classification framework that directly predicts all mutated genes from the input WSIs. BPGT consists of three modules: the visual extractor, the gene encoder, and the label decoder, among which the latter two are our newly designed modules that utilize comprehensive biological gene priors to enhance the mutation classification accuracy. The overall workflow is previewed as follows:
% The overall workflow and the designed modules are previewed as follows:

% (1) The overall workflow: 
Our BPGT firstly designs a \textbf{visual extractor} (VE, Fig. \ref{fig_2} (a)) that slices the input WSI (denoted as $\mathbf{X}$) into numerous patches and extracts patch-level histopathology features (denoted as $\mathbf{F}$). BPGT simultaneously designs a graph-based \textbf{gene encoder} (GE, Fig. \ref{fig_2} (b)). GE firstly constructs a gene graph (GG, Fig. \ref{fig_2} (b1)) by encoding the genes' linguistic knowledge as initial node features $\mathbf{G}$ and their biomedical knowledge as edges and node weights. GE then designs a knowledge association module (KAM, Fig. \ref{fig_2} (b2)) to aggregate the linguistic and biomedical knowledge to obtain the gene priors $\mathbf{P}$, which considers the mutation correlations between different genes. Next, BPGT feeds visual features $\mathbf{F}$ and the gene priors $\mathbf{P}$ to the \textbf{label decoder} (LD, Fig. \ref{fig_2} (c)), which designs a modality fusion module (MFM, Fig. \ref{fig_2} (c1)) to integrate the visual features and gene priors, guiding the multi-label classifier to focus on critical WSI parts concerning the genetic mutations. The predicted gene mutation logits $\mathbf{y}$ are supervised by the slide-level mutation labels $\mathbf{y}^*$ using a comparative multi-label loss (Fig. \ref{fig_2} (c2)) to better discriminate the mutated genes from the non-mutated ones.

\subsection{Visual extractor}  % $\mathbbm{v}$
Our BPGT first adopts a VE to effectively extract patch-level semantic features from the gigapixel WSIs. 
Following the widely-used preprocessing procedure\cite{lu2021data}, the VE first divides the input WSI into non-overlapping patches. Meanwhile, the background area (white-colored areas in Fig. \ref{fig_2} (a)) of the WSI is determined by the OTSU algorithm; the WSI patches containing background area (patches out of the green contours in Fig. \ref{fig_2} (a)) are then removed to reduce computational cost.
Then, a segmentation algorithm is utilized to further remove the patches with the cavity (patches in the blue contours shown in Fig. \ref{fig_2} (a)). In this way, the whole image is sliced into multiple image patches containing tissues that are hematoxylin and eosin (H$\&$E) stained in Fig. \ref{fig_2} (a). These patches are all of the size $L\times L$, which are denoted as $\mathbf{X}={\mathbf{x_1},\dots,\mathbf{x}_{N}} \in \mathbb{R}^{N \times L \times L \times 3}$, where $\mathbf{x_i}$ represents i-th WSI patch.
Next, a pre-trained feature extractor (DINO network \cite{chen2022self}) is employed to extract coarse-grained patch-level features. These features are then flattened and fed into the patch-level visual encoder, which outputs fine-grained visual features $\mathbf{F} \in \mathbb{R}^{N \times d}$ for downstream slide-level tasks.
To verify that our BPGT generally achieves high performances on different visual encoders, we select some SOTA visual encoders detailed as follows:
\begin{itemize}
    \item GeneHe-VE \cite{qu2021genetic}: After extracting the coarse-grained patch-level features, GeneHe-VE first selects the patches containing cancer tissues by a rough K-means classification. It then adopts the scaled dot-product multi-head self-attention layer in classic transformer \cite{vaswani2017attention} for information interaction among patches, which obtains fine-grained patch-level visual features $\mathbf{F}$.
    \item DeepHis-VE \cite{kather2020pan}: The main workflow of DeepHis-VE is the same as GeneHe-VE. However, it randomly selects 500 patches (not necessarily containing tumor tissues) from $\mathbf{X}$. Also, it simply utilizes a 3-layer MLP to extract the fine-grained patch-level features for each patch as $\mathbf{F}$. 
    \item Attention-VE (Att-VE)\cite{ilse2018attention}: The main workflow of Att-VE is also similar to GeneHe-VE, however, it does not perform patch selection. Instead, Att-VE feeds the coarse-grained patch-level features from all patches in $\mathbf{X}$ to several multi-head self-attention layers \cite{zhang2019self} to calculate a score for each patch. Next, Att-VE carries out feature interaction among patches weighted by the scores to calculate $\mathbf{F}$.
    \item Transformer-VE (Trans-VE)\cite{shao2021transmil}: Similar to Att-VE, Trans-VE feeds the coarse-grained patch-level features of all patches to a CNN to obtain their positional encodings, which are then concatenated with the patch features and fed to a vision transformer (ViT) \cite{dosovitskiy2020image} to calculate the attention scores between patches and perform patch-level feature interaction. The interacted features are used as the $\mathbf{F}$. 
    \item Kernel Attention Transformer (KAT) \cite{zheng2022kernel}: Similar to Trans-VE, KAT also uses the ViT attention mechanism for feature interaction, however, it calculates a mask matrix encoding the spatial distances of the patches, which is multiplied by the value matrix when performing the patch-level feature interaction. In this way, the feature interaction among patches is guided by their spatial distance to extract the fine-grained patch-level visual features $\mathbf{F}$.
    
    \item Hierarchical Image Pyramid Transformer (HIPT) \cite{chen2022scaling}: It first reorganizes the coarse-grained patch-level features to align with the spatial arrangement of the larger divided patches (e.g., each set of $16 \times 16$ features from patches with the size of $256 \times 256$ are concatenated to form a new feature for each $4096 \times 4096$ patch). Subsequently, these rearranged features are input into another Dino to perform feature interaction among patches, yielding outputs as the fine-grained features $\mathbf{F}$. 

\end{itemize}
\vspace{0.1cm}
\begin{figure*}[htbp]
	\centering
	\includegraphics[width=0.9\linewidth]{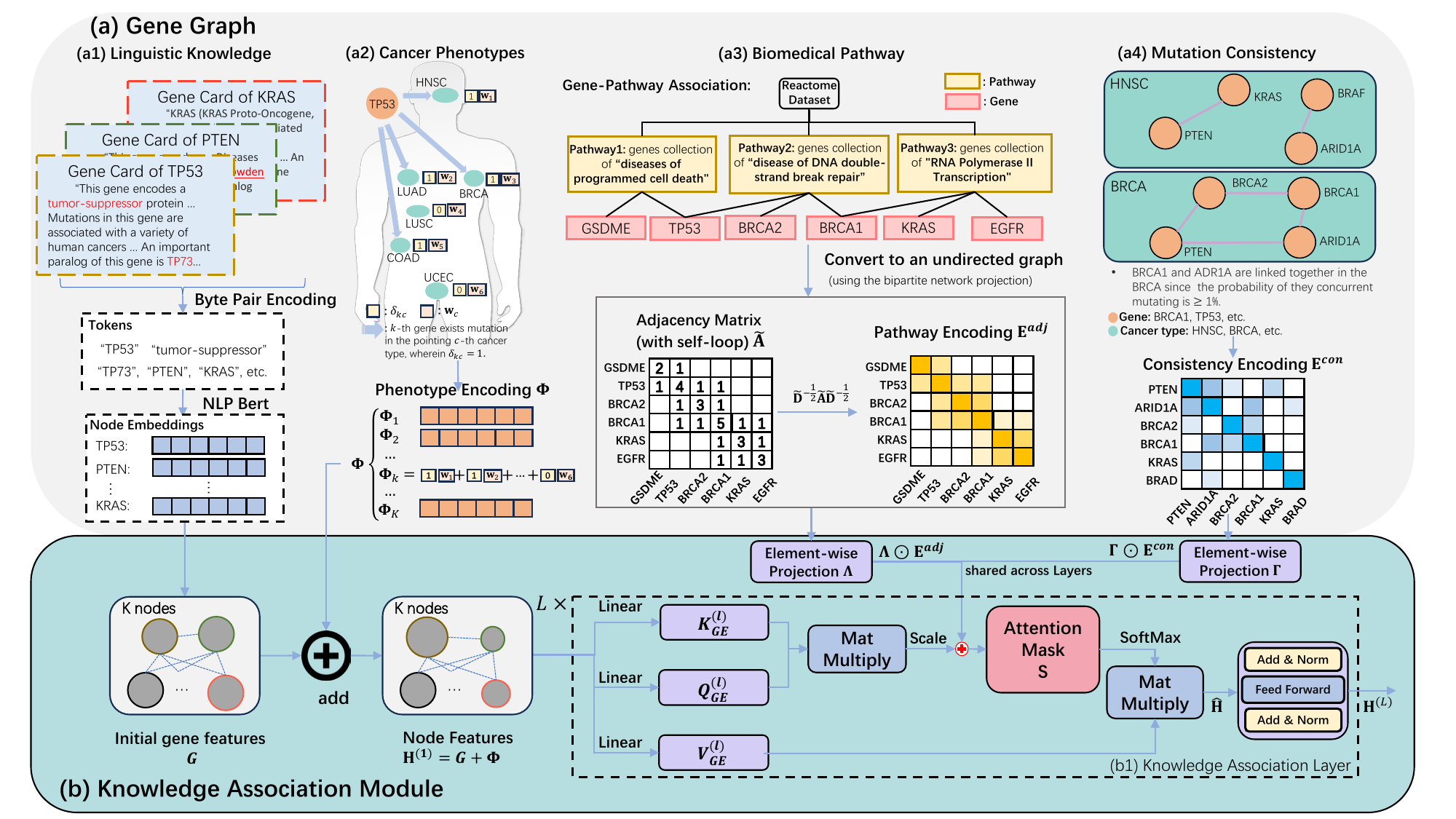}
	\caption{Gene encoder (GE) is designed to aggregate linguistic and biomedical knowledge into gene priors. GE contains a gene graph (GG, Fig. \ref{fig_3} (a)) and a knowledge association module (KAM, Fig. \ref{fig_3} (b)). GG considers linguistic knowledge and biomedical knowledge encoding. Linguistic knowledge encoding (Fig. \ref{fig_3} (a1)) is firstly obtained from the GeneCard and encoded via byte pair encoding and the NLP Bert, which is utilized as the initial gene features. Biomedical knowledge encoding contains three encoding approaches: Phenotype encoding (Fig. \ref{fig_3} (a2)) encodes cancer types for leveraging the gene-cancer relationships to help predict gene mutation; pathway encoding (Fig. \ref{fig_3} (a3)) encodes the biomedical functions of different genes to consider their mutation relationship; consistency encoding (Fig. \ref{fig_3} (a4)) encodes the concurrent mutation frequency of different genes from the data-driven aspect. KAM designs transformer-based graph representation learning, which introduces the linguistic and phenotype encoding in node features, and the pathway and consistency encoding in edge weights; KAM thereby integrates the above four types of genetic knowledge into gene priors.} 
	\label{fig_3} 
	 % \vspace{-0.5cm}
\end{figure*}
\subsection{Gene encoder}\label{mtd:intro}
GE is designed to establish the gene priors $\mathbf{P}$ via constructing a gene graph (GG, Fig. \ref{fig_2} (b1)) and designing knowledge association modules (KAM, Fig. \ref{fig_2} (b2)) to integrate the biomedical and linguistic knowledge in the gene labels, which helps to explore the relationships of the mutations between genes.
\subsubsection{Gene graph}\label{mtd:main:geg}
As shown in Fig. \ref{fig_2} (b1), GG encodes the genes' linguistic knowledge to establish the initial node features (i.e., the initial gene features) $\mathbf{G}$; it also considers the genes' biomedical knowledge to build the edge and node weights.

\noindent\textbf{Linguistic knowledge encoding.}
The initial gene features $\mathbf{G}$ are designed to encode the linguistic knowledge behind genes to capture unique information about each label.
% , which also reflects the genes' distances in the feature space. 
The linguistic knowledge is collected from the GeneCard \footnote{\href{https://www.genecards.org/}{https://www.genecards.org}}. This knowledge contains each gene's linguistic biological description, including information about mutation characteristics, biological functions, clinical significance, etc. For example, the linguistic knowledge of the TP53 gene is: "This gene encodes a tumor-suppressor protein ... Mutations in this gene are associated with a variety of human cancers ... An important paralog of this gene is TP73...". This knowledge indicates that the TP53 gene encodes a tumor suppressor protein; thus, its mutation will increase predisposition to cancer. Furthermore, TP53 may share common biological functions with other tumor suppressor genes such as TP73 and TP63, however, they may possess opposite functions from tumor promoter genes (oncogenes) such as KRAS, and EGFR. Thus, by incorporating this linguistic knowledge in the initial node features, we can comprehensively capture the uniqueness of each gene and its linguistic associations with other genes, which are beneficial for capturing the relationships between the mutation of different genes and that between genetic mutation and cancer occurrence. As a comparison, simply using one-hot vectors to represent each gene may cause an over-smooth problem in graph learning~\cite{zheng2022rethinking}. Furthermore, this initialization does not contain genetic linguistic information, which would significantly weaken the performance of the successive graph representation learning.  
We adopt Byte Pair Encoding (BPE) ~\cite{shibata1999byte} to tokenize this textual linguistic knowledge from the GeneCard into sub-word tokens.  
The tokens can preserve key concepts in the linguistic knowledge such as the gene names ``TP53/TP73'' and the biological function name ``tumor-suppressor''. 
These tokens are then fed into pre-trained BERT models to extract gene features $\mathbf{G}$, which are utilized as the initial node features. The BERTs we employ have been pre-trained on relevant biological corpora, which enables the gene features $\mathbf{G}$ to capture the inherent semantic distinctiveness and relevance of the genes. For instance, Bio-BERT \cite{lee2020biobert} is trained using large-scale corpora encompassing biological text with gene names, functions, and relations. Thus, the gene features $\mathbf{G}$ obtained from Bio-BERT could reflect the semantic uniqueness and relevance embedded within their linguistic knowledge.

\noindent\textbf{Biomedical knowledge encoding.}
Having calculated the initial node features $\mathbf{G}$ for each node (gene), GE further considers the intrinsic biomedical knowledge within labels by designing three encoding approaches: 
(1) Phenotype encoding $\mathbf{\Phi}$ that encodes cancer types for leveraging the gene-cancer relationships to help predict gene mutation. 
(2) Pathway encoding $\mathbf{E}^{adj}$ that encodes the biomedical functions of different genes to consider their mutation relationships from the biomedical aspect. 
(3) Consistency encoding $\mathbf{E}^{con}$ that encodes the concurrent mutation frequency of different genes from the data-driven aspect. 
The encoded features would reflect the intrinsic biomedical knowledge (e.g., the mutation relationships), which will be leveraged by the knowledge association layer (Section 2.2.2) to construct the gene priors to reflect the dependencies among different genes.

\noindent\textit{\textbf{-Phenotype encoding.}} Phenotype encoding leverages the gene-cancer relationships to encode the association of gene mutations with cancers, which is introduced into node features to improve their discrimination. 
Phenotype encoding benefits genetic mutation prediction because different cancers may be associated with the mutation of different genes. 
Intuitively, phenotype encoding statistically analyzes the genes' associations with cancers, i.e., if statistical information indicates that a gene is mutated in many types of cancer, then this gene is considered to be more likely to mutate in tumorous WSIs. Thus, our phenotype encoding introduces this intuition by considering the potential gene-cancer relationships and adding a bias term to the node features to indicate this statistical information.
For example, as shown in Fig. \ref{fig_3} (a2), the mutation of gene TP53 occurs in multiple cancers such as breast cancer (BRCA), lung cancer (LUAD), colon cancer (COAD), and head and neck cancer (HNSC); the learnable features of these four types of cancers will thus be added to the phenotype encoding of the TP53 gene. More formally, our phenotype encoding is formulated as a phenotype matrix $\mathbf{\Phi} \in \mathbb{R}^{K \times d}$, where $K$ represents the number of genes and $d$ represents the dimension of node features (the same as those of the gene features $\mathbf{G}$). The \textit{k}-th row in $\mathbf{\Phi}$ (i.e., $\mathbf{\phi}_k$) means the phenotype encoding of the \textit{k}-th gene, i.e.,:
\begin{equation}
	\mathbf{\phi}_k  = \sum^{C}_{c} \delta_{kc} \mathbf{w}_{c} 
\end{equation}
where $\delta_{kc}$ indicates whether the mutation of the $k$-th gene is associated with the occurrence of the $c$-th cancer type (1 for associated and 0 for not associated), which is obtained based on empirical genetic mutation patterns in human cancers~\cite{alexandrov2020repertoire}. 
The vector $\mathbf{w}_{c}$ is a learnable parameter that represents the $c$-th cancer, which serves a similar purpose as the learnable positional embeddings in the transformer~\cite{parmar2018image}.
Therefore, when we compute the phenotype encoding $\mathbf{\phi}_k$, the features of cancers that are associated with the mutation of the $k$-th gene are aggregated in $\mathbf{\phi}_k$.

\noindent\textit{\textbf{-Pathway encoding.}} Pathway encoding leverages the biomedical pathways, i.e., the collections of functionally related genes \cite{ramanan2012pathway}, to exploit the genes' intrinsic functional relationships. 
Pathway encoding benefits genetic mutation prediction by leveraging these functional relationships as the edges to link the related genes in the GG, thereby increasing the predicted probability of their concurrent mutation to align with these relationships.
Pathway encoding first extracts gene-pathway relationships from the Reactome database \cite{gillespie2022reactome}, which could be constructed as a two-layer tree structure from the pathways to the genes as shown in Fig. \ref{fig_3} (a3). Each layer in the tree structure is a set, i.e., the pathways and the genes can be regarded as two sets (the pathway set is illustrated by the yellow-filled boxes in Fig. \ref{fig_3} (a3), whereas the gene set is illustrated by the red-filled boxes in Fig. \ref{fig_3} (a3)). The connections between the sets in Fig. \ref{fig_3} (a3) show the genes' functional relationships, for example, the pathway named "diseases of DNA double-strand break repair" is a collection of BRCA1, BRCA2, and TP53 genes, which means that this disease could be caused by the mutation in these functionally related genes.  
Pathway encoding then adopts the bipartite network projection~\cite{zhou2007bipartite} to convert the tree structure into an undirected graph. For example, for the tree structure in Fig. \ref{fig_3} (a3), the bipartite network projection uses one set (i.e., the gene set) as nodes and converts their shared connections with the other set into edges (i.e., the gene set's connections to the pathway set are transformed to the edges). For a more detailed example, the gene ``TP53'' in the gene set is converted to a node, whereas its connections with the pathway set are converted into the edges connected to this node (i.e., the genes GSDME, BRCA2, and BRCA1 are connected to TP53 by Pathway 1 and Pathway2, thus, the corresponding edge weights $\tilde{A}_{21}$/$\tilde{A}_{12}$, $\tilde{A}_{23}$/$\tilde{A}_{32}$, and $\tilde{A}_{24}$/$\tilde{A}_{42}$ in Fig. \ref{fig_3} (a3) are increased by 1). 
More formally, in the converted graph, the nodes represent genes, and the edges are represented by an adjacency matrix (with self-loop), which are denoted as $\tilde{\mathbf{A}}\in \mathbb{R}^{K \times K}$. Each element $\tilde{A}_{ij}$ in the matrix are calculated by:
\begin{equation}
	   \tilde{A}_{ij} = 
	\begin{cases}
	   	max(n_{ij|j=\{1:K\}\setminus \{i\}})+1 &  \text{if}  \quad i = j \\
		n_{ij} &  \text{else} 
	\end{cases}
\end{equation}
where $n_{ji}$ denotes the number of pathways connected to both gene $i$ and gene $j$, which represents their functional similarities. For example, since the second and fourth gene in Fig. \ref{fig_3} (a3) (TP53 and BRCA1) are both connected to one pathway (Pathway2), the elements $\tilde{A}_{24}$ and $\tilde{A}_{42}$ would be both 1. 
Pathway encoding next normalizes $\tilde{\mathbf{A}}$ as in classical GCN\cite{kipf2016semi}, which yields the pathway encoding matrix $\mathbf{E}^{adj}=\tilde{\mathbf{D}}^{-\frac{1}{2}}\tilde{\mathbf{A}}\tilde{\mathbf{D}}^{-\frac{1}{2}}$, where $\tilde{\mathbf{D}}$ is a diagonal matrix and $\tilde{D}_{ii} = \sum_j \tilde{A}_{ij}$. 
In this way, $\mathbf{E}^{adj}$ helps our BPGT to be aware of the genetic mutation relationships from the aspect of biomedical functions.

\noindent\textit{\textbf{-Consistency encoding.}}
Having considered the genetic mutation relationships from the aspect of biomedical functions, another inspiration is to consider the mutation relationships from the aspect of statistical data information. Consistency encoding is thus designed to quantify the gene concurrent mutation frequency in the dataset, which is also leveraged as edges in the GG to achieve the same goal as pathway encoding. Consistency encoding calculates the probability of any two genes having a concurrent mutation in all WSIs in the dataset for each cancer; if the probability of the two genes having a concurrent mutation is greater than 1\%, the two genes are considered to be statistically correlated for mutation.
More formally, the consistency encoding $\mathbf{E}^{con} \in \mathbb{R}^{K \times K}$ is defined as follows:  
\begin{equation}
	E^{con}_{ij}  = 
	\begin{cases}
		1 &  \text{if}  \quad i = j \\  
		\frac{1}{N} \sum^C_{c} \zeta_{ijc} N_c  &  \text{else}
	\end{cases}
\end{equation}
where $N$ is the number of WSIs in the training set, and $N_c$ is the number of WSIs of the $c$-th cancer type. $\zeta_{ijc}$ is a binary variable; $\zeta_{ijc}=1$ if both $i$-th and $j$-th genes exhibit a mutation with at least 1$\%$ mutation probability in the $c$-th cancer type (this 1\% is a relatively large threshold because the gene mutation concurrency is relatively low in the huge number of genes). The weighted average among all cancer types means considering all tumorous WSIs in terms of their dataset scale. We have also attempted another setting that decides the value of $E^{con}_{ij}$ based on the probability of concurrence (rather than setting it to 1 if it exceeds 1$\%$ as we now do). However, this approach does not yield significant improvements, which could be caused by the bias introduced by the discrepancy in the distribution of concurrent mutation probabilities between the training and testing set. In contrast, the consistency encoding matrix with the ``truncate'' setting may be better for capturing the statistical genetic mutation concurrence relationships.

\vspace{0.1cm}
\begin{figure*}[htbp]
	\centering
	\includegraphics[width=0.95\linewidth]{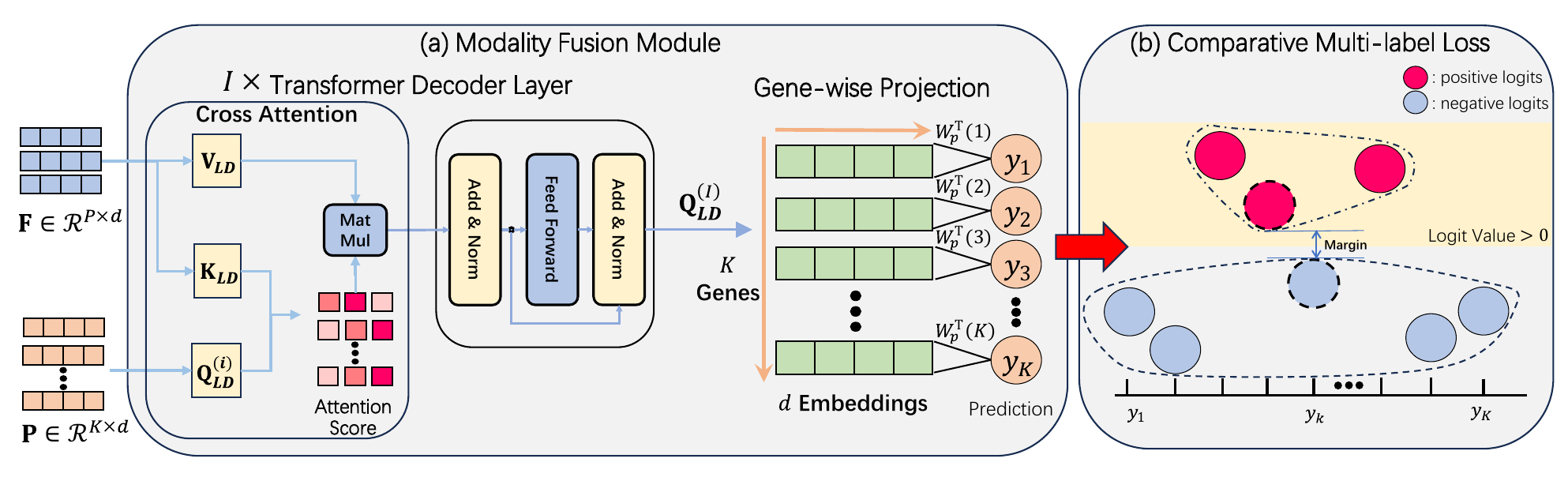}
	\caption{Label decoder (LD) is designed to integrate the gene priors $\mathbf{P}$ and the visual features $\mathbf{F}$, which enables the gene priors in $\mathbf{P}$ to guide the model for the multi-label classification. (a) The modality fusion module firstly leverages transformer decoder layers to integrate visual features with the gene priors to obtain embeddings $\mathbf{Q}^{(I)}_{LD}$ via a cross-attention mechanism. Then, the gene-wise projection independently maps each row of $\mathbf{Q}^{(I)}_{LD}$ to its corresponding gene prediction (logit) by multiplying it with a unique learnable column vector. (b) Multi-label loss is designed to enlarge the margin between the hardest positive and negative logits (red and blue circles with dot lines in Fig.~\ref{fig_11} (b)) to increase the discrimination for positive and negative predictions, wherein the positive and negative prediction is determined based on logit value $>0$ or $\leq 0$.} 
	\label{fig_11} 
	 % \vspace{-0.5cm}
\end{figure*}

\subsubsection{Knowledge association module.} \label{mtd:main:kal}
The KAM (Fig. \ref{fig_3} (b)) is designed to effectively evolve the node features in GG with the help of biomedical knowledge encodings in a transformer-based graph representation learning manner. KAM constructs the gene prior $\mathbf{P}$ for each gene that comprehensively encodes linguistic knowledge and biomedical relationships. This helps capture the relationships and dependencies among the gene mutations, which enhances the model's ability for multi-label classification. 
More formally, as illustrated in Fig. \ref{fig_3} (b), our KAM is inspired by the transformer encoder layer \cite{vaswani2017attention} that constructs an attention mask and utilizes the attention mechanism to implement graph representation learning. KAM first adds the phenotype encoding $\mathbf{\Phi}$ to the initial gene features $\mathbf{G}$:
\begin{equation}
        \mathbf{H}^{(1)} = \mathbf{G} + \mathbf{\Phi}    
\end{equation}
In this way, $\mathbf{H}^{(1)}$ integrates each gene's linguistic knowledge with its associated cancer features, which better reflects the inherent characteristics of each gene. 
Then, the $\mathbf{H}^{(1)}$ is fed into several knowledge association layers (KAL, Fig. \ref{fig_3} (b1)) to consider the effect of the pathway and consistency encoding as bias terms for the feature evolution. In each KAL (take the $l$-th layer as an example), an attention mask $\mathbf{S}^{(l)}$ is constructed using the query-key production manner:
\begin{equation}\label{eq10}
    \mathbf{S}^{(l)} = \frac{\mathbf{Q}^{(l)}_{GE}(\mathbf{K}^{(l)}_{GE})^{\intercal}}{\sqrt{d}} + \mathbf{\Gamma} \odot \mathbf{E}^{con} + \mathbf{\Lambda} \odot \mathbf{E}^{adj}  % 改成矩阵形式的公式
\end{equation}
where $\mathbf{Q}^{(l)}_{GE}$ denotes the query matrix calculated by $\mathbf{Q}^{(l)}_{GE}=\mathbf{H}^{(l)}\mathbf{W}^{(l)}_{Q}$ (the subscript GE means the matrices in the graph encoder), $\mathbf{K}^{(l)}_{GE}$ denotes the key matrix calculated by $\mathbf{K}^{(l)}_{GE}=\mathbf{H}^{(l)}\mathbf{W}^{(l)}_{K}$, $\mathbf{H}^{(l)}$ denotes the $l$-th layer's input, $\mathbf{W}^{(l)}_{Q}$ and $\mathbf{W}^{(l)}_{K}$ are learnable matrices. $\mathbf{\Gamma}$ and $\mathbf{\Lambda}$ are also learnable matrices, whose values are shared across all knowledge association layers. Subsequently, the node features are evolved using standard self-attention:
\begin{equation}\label{eq11}
    \hat{\mathbf{H}}^{(l+1)} = softmax(\mathbf{S}^{(l)})\mathbf{V}^{(l)}_{GE} ,
\end{equation}
where $\mathbf{V}^{(l)}_{GE}$ denotes the value matrix calculated by $\mathbf{V}^{(l)}_{GE}=\mathbf{H}^{(l)}\mathbf{W}^{(l)}_{V}$, $\mathbf{W}^{(l))}_V$ is also a trainable matrix. Then, $\hat{\mathbf{H}}^{(l+1)}$ are fed into three consecutive layers (an Add\&Norm layer, a feed-forward layer implemented by an MLP, and another Add\&Norm layer) as in classical transformer \cite{vaswani2017attention}, which yields $\mathbf{H}^{(l+1)}$, the output of the \textit{l}-th KAL. After \textit{L} KAL layers, the KAM outputs the gene priors \textbf{P} for all genes in the GG.

The gene priors \textbf{P} constructed by KAM comprehensively consider the linguistic and biomedical knowledge relationships regarding the genetic mutation correlations. The attention mask $\mathbf{S}$ is calculated by query-key production, which considers the gene's inherent characteristics (i.e., the linguistic knowledge and the gene-cancer phenotype association). Furthermore, adding the consistency encoding $\mathbf{E}^{con}$ and the pathway encoding $\mathbf{E}^{adj}$ as bias terms to $\mathbf{S}$ considers the relationships of genetic mutation concurrence derived from pathways and consistency. In this way, the aggregation of node features in KAM considers the comprehensive genetic prior knowledge, which is beneficial for capturing the mutation relationships among different genes in the multi-label classification paradigm.

\subsection{Label decoder}
LD designs a modality fusion module containing several sequential transformer decoder layers and a label-wise projection layer to link the gene priors with the gigapixel WSI features, which helps our BPGT to focus on critical WSI parts; it then designs a comparative multi-label loss to achieve genetic mutation classification, which helps BPGT better find all mutated genes in the multi-label classification paradigm. 
\subsubsection{Modality fusion module}
As shown in Fig. \ref{fig_2} (c), our MFM contains several transformer decoder layers and a gene-wise projection layer. In each transformer decoder layer, the fine-grained patch-level visual features $\mathbf{F}$ (obtained from the visual extractor) will be projected as key and value matrices, which are respectively denoted as $\mathbf{K}_{LD}=\mathbf{F}\mathbf{M}_{K}$ and $\mathbf{V}_{LD}=\mathbf{F}\mathbf{M}_{V}$, which are shared among different transformer decoder layers. 
Accordingly, the gene prior $\mathbf{P}$ (obtained by the gene encoder) is leveraged as the query matrix $\mathbf{Q}^{(1)}_{LD}$ for the first transformer decoder layer.  % i
For the $i$-th transformer decoder layer, the output of its cross-attention layer is computed as follows:
\begin{equation}
    \hat{\mathbf{Q}}^{(i+1)}_{LD}  = LN(softmax( \frac{\mathbf{Q}^{(i)}_{LD}\mathbf{K}^{\intercal}_{LD}}{\sqrt{d}}) \mathbf{V}_{LD} + \mathbf{Q}^{(i)}_{LD}) \label{eq1}.
\end{equation}
where $d$ is the column dimension of $\hat{\mathbf{Q}}^{(i)}_{LD}$, and $\textit{LN}$ represents the layer normalization. In this manner, Eq.~\ref{eq1} incorporates the visual features $\mathbf{F}$ from WSIs with the gene prior $\mathbf{P}$ via the cross-attention mechanism. Then, a skip connection and a feed-forward layer $\textit{FFN}$ are utilized to yield the output of the transformer decoder layer, which is formulated as $\mathbf{Q}^{(i+1)}_{LD}=LN(FFN(\hat{\mathbf{Q}}^{(i+1)}_{LD})+\hat{\mathbf{Q}}^{(i+1)}_{LD})$. $\mathbf{Q}^{(i+1)}_{LD}$ is iteratively updated by each layer (i.e., the first decoder layer takes $\mathbf{Q}^{(1)}_{LD}$ as the query while the rest takes the $\mathbf{Q}^{(i)}_{LD}$ from its previous layer as the query). 
The last ($I$-th) transformer decoder layer's output is $\mathbf{Q}^{(I)}_{LD} \in \mathbb{R}^{K\times d}$. Next, we use a gene-wise projection to predict the final mutation logits of all \textit{K} genes. In this procedure, \textit{K} learnable vectors of dimension $d \times 1$ are used, and each row of the $\mathbf{Q}^{(I)}_{LD}$ (representing the features of one gene) is multiplied with one learnable vector to produce a scalar that represents the mutation score of this gene. The learnable vectors multiplied to different rows (genes) are not the same, i.e., each vector is only responsible for learning the mutation state for one gene. In this way, the gene-wise projection produces the logits $\mathbf{y} \in \mathbb{R}^{K}$ indicating the mutation score of all genes, which will be used to calculate the comparative multi-label loss in Section 2.3.2. For the final prediction, a sigmoid function is applied to the $\mathbf{y}$ to obtain the final mutation probability vector ranging from 0 to 1 (each element in the vector represents the mutation probability of each gene). For a given threshold $\beta$, (e.g., 0.5), if an element in the mutation probability vector is larger than $\beta$, then this gene is predicted to be mutated and vice versa.

\subsubsection{Comparative multi-label loss}
The comparative multi-label loss is designed to better distinguish all mutated (positive) genes from the non-mutated (negative) ones. Since there can be more than one mutated gene in one sample (WSI), directly using the softmax cross-entropy loss in the classic classification task is infeasible (because it allows only one class as "positive" for each sample, i.e., it can only predict one gene as mutated for each WSI). 
This is why popular MIL methods individually train multiple binary classifiers for all genes. However, as mentioned in the Introduction section, this strategy requires training hundreds of independent binary classifiers, which has the drawbacks:
(1) This strategy inevitably gives rise to a class imbalance problem. When predicting the mutation status for each gene, the number of samples with non-mutated states always greatly exceeds the number of samples with mutated states. 
(2) The binary classifiers independently calculated for each class ignore the relative comparisons among classes. 
In contrast, the softmax loss not only avoids introducing the imbalance issue but also inherently considers comparisons among classes\cite{sun2020circle}. 
The comparative multi-label loss is thus inspired to leverage this ability of softmax loss to better distinguish all mutated genes from the non-mutated ones in the multi-label classification paradigm. Also, we add a margin in the loss to improve the discrimination of the hardest positive (i.e., the gene is mutated but with the lowest predicted score in all positive classes) and negative class (vice versa): 
\begin{align}
    \mathcal{L} = softplus\left[\log{\sum\limits_{n\in\mathcal{N}}e^{y_n}} + T\log{\sum\limits_{p\in\mathcal{P}}e^{-\frac{y_p}{T}}}\right]\label{eq:4}
     % & = \log \left(1 + \sum\limits_{n\in\mathcal{N},p\in\mathcal{P}}e^{y_n-y_p}\right) \qquad w.r.t. \quad T=1 \label{eq:5}
\end{align}
where $softplus(\cdot) = \log[1 + \exp(\cdot)]$; the sets $\mathcal{N}$ and $\mathcal{P}$ respectively represent the classes that are actually negative (non-mutated) and positive (mutated) in $y^*$; the $y_{n}$ and $y_{p}$ indicate the predicted logits for the positive and negative classes; the summations $\sum\limits_{n\in\mathcal{N}}$ and $\sum\limits_{n\in\mathcal{P}}$ mean respectively summing up the exponential logits of all negative and positive classes. $T$ controls the scale of the margin between the hardest positive and negative logits. % 

Eq.~\ref{eq:4} inherently compares among classes in softmax for prompting discrimination. This is because Eq.~\ref{eq:4} is derived from the softmax loss by replacing the single positive logit from the softmax loss with the hardest logit among all positive logits and additionally introducing a margin.  
The log-sum-exp in Eq.~\ref{eq:4} resembles a maximum operator, thus, the first part of Eq.~\ref{eq:4} is equivalent to finding the negative class with the maximum predicted logit, while the second part finds the positive class with the minimum predicted logit, i.e., the hardest negative/positive classes. As a result, the loss function aims to enlarge the logit margin between the hardest positive and negative classes, i.e., it forces the lowest positive logit to be higher than the highest negative logit by some margin (controlled by the hyper-parameter T ($T>0$)), which effectively improves the BPGT's discrimination for positive and negative classes. 
Additionally, Eq.~\ref{eq:4} could inherently alleviate the imbalance issue in MIL methods because it compares the mutation probabilities among different genes, instead of independently comparing the probabilities of a gene's mutation status and non-mutation status.

%%%%%%%%%%%%%
% Experiment
%%%%%%%%%%%%%
\section{Experiments}
\subsection{Experimental setup}
\subsubsection{Data preparation and implementation details}
In this study, we evaluate the genetic mutation prediction performance of BPGT using a challenging dataset (The Cancer Genome Atlas, TCGA) \cite{kandoth2013mutational}.
TCGA is challenging for genetic mutation prediction because: (1) TCGA contains WSIs of large sizes (the widths and heights of the WSIs are typically 50$\sim$900 thousand pixels) with only patient-level genetic mutation labels, which makes it difficult to locate the regions related to the mutation. (2) TCGA is a comprehensive dataset containing genomic mutation information from various cancer types, where there may be more than one mutated gene for each WSI. Thus, TCGA is chosen for assessing our BPGT's ability to accurately identify genetic mutations associated with different types of cancer. 
To ensure the statistical significance of genetic mutations and cancers, we select the top 9 cancers with the highest incidence rates; we also select genes with mutation frequencies above 1$\%$ in each cancer as suggested in~\cite{qu2021genetic}. These procedures construct a dataset containing 3,800 WSI slides and 30 genes. This dataset is split into the training set and the test set using standard five-fold cross-validation. 

BPGT is implemented by PyTorch based on Python 3.6.5 and trained using the Adam optimizer with a learning rate of $10^{-4}$. The other hyper-parameters are set as in \cite{zhang2018improved}, i.e., $\beta_1=0.9$, $\beta_2=0.999$ and $\epsilon=10^{-8}$. All the trainable parameters are initialized with the Xavier method. BPGT is trained for 250 epochs. We use a $L_2$ regularizer, and the L2 loss weight is $10^{-5}$. All evaluation metrics (detailed in Section 3.1.2) are calculated by averaging performances on the 5-fold cross-validation.

\subsubsection{Evaluation metrics}
To evaluate the genetic mutation prediction performance for each gene, we follow~\cite{qu2021genetic, kather2020pan, chen2023optimization, guo2022robust} to adopt the per-class F1 score and per-class AUC, where each class indicates each gene. We further adopt the overall F1 score and overall AUC for all different genes, which comprehensively measure the performance of the overall multi-label classification. 

\noindent\textbf{Per-class AUC and overall AUC.} %
The per-class AUC evaluates the sensitivity and specificity of the mutation classification of each gene. Per-class AUC is the area under the receiver operating characteristic (ROC) curve, which is a graphical plot of the true positive rate (TPR) against the false positive rate (FPR) for each gene's mutation status classification at various threshold settings $\beta$ (the cut-off value used by the classifier to distinguish whether the gene is mutated or not mutated). As the threshold $\beta$ varies, a set of TPR and FPR are defined as:
\begin{equation}
\text{TPR}(\beta)=\frac{\text{TP}(\beta)}{\text{TP}(\beta)+\text{FN}(\beta)}, \quad \text{FPR}(\beta)=\frac{\text{FP}(\beta)}{\text{FP}(\beta)+\text{TN}(\beta)} \nonumber
\end{equation}
where $\text{TP}(\beta)$, $\text{TN}(\beta)$, $\text{FP}(\beta)$, $\text{FN}(\beta)$ represent the true positive (TP), true negative (TN), false positive (FP), and false negative (FN) values for the threshold $\beta$. By considering all TPRs and FPRs at different thresholds ($\beta$ values) as points in a two-dimensional space, we can construct the TPR-FPR curve, commonly known as the ROC curve. The AUC is obtained by computing the area under this curve. Then, The overall AUC is obtained by averaging the per-class AUC values calculated for all genes.

\noindent\textbf{Per-class and overall F1 score.}
The per-class and overall F1 score are metrics that can provide a comprehensive evaluation of the performance of multi-label classification at a specific threshold $\beta$ (e.g., 0.5 as recommended in \cite{zhu2017learning}). For the per-class F1 score, the F1 score of the $i$-th class is defined as $F1_{i} = \frac{2 \times P_{i} \times R_{i}}{P_{i}+R_{i}}$, wherein $P_{i}$ and $R_{i}$ represent the per-class precision and recall. They are respectively calculated as $P_{i}=\frac{TP_{i}}{TP_{i}+FP_{i}}$ and $R_{i}=\frac{TP_{i}}{TP_{i}+FN_{i}}$, where $i$ represents the $i$-th label. 
The overall precision, recall, and F1 score (OP, OR, OF1) are defined as follows:
\begin{equation}
    OP = \frac{\sum_{i}TP_{i}}{\sum_{i}P_{i}},\quad
    OR = \frac{\sum_{i}TP_{i}}{\sum_{i}P_{i}^{*}}, \quad
    OF1 =  \frac{2 \times OP \times OR}{OP+OR}  \nonumber
\end{equation}
where $P_{i}=TP_{i}+FP_{i}$ and $P_{i}^{*}=TP_{i}+FN_{i}$.

\subsection{Overall genetic mutation prediction performance}
The versatility of BPGT is demonstrated by its high performance across different visual extractors, as shown in Table. \ref{tab12}. Each row of Table. ~\ref{tab12} represents the gene mutation classification results by using the different VE backbones discussed in Section 2.1; each column corresponds to the predictive performance for a specific gene, measured in terms of the AUC and F1 scores, while the last column reflects the overall performance across all genes. Notably, for the Trans-VE backbone, the mutation prediction performance for TP53 genes could achieve 74.5$\%$ AUC and 69.6$\%$ F1, whereas the Att-VE and HIPT backbones could also obtain high AUC and F1 scores for various genes, such as 72.3$\%$ AUC and 64.9$\%$ F1 on TP53 and 73.3$\%$ AUC and 69.1$\%$ F1 on PTEN. Similarly, regarding the overall gene mutation performance, we observed high values for OF1 could also be achieved across different VE backbones. These findings validate the effectiveness of our BPGT model when applied to different VE backbones.
\begin{table*}[ht]
    % \sisetup{table-format=2.2}
    \caption{BPGT's high genetic mutation prediction performance with varying visual extractor backbones. The '-VE' notation represents constructing BPGT using the visual extractor of the reference methods with the GE and LD. Per-class and overall AUC/F1 scores are presented to demonstrate the performance.}\label{tab12}
    \centering \small
    % \footnotesize  
    \resizebox{\textwidth}{!}{
    \begin{tabular}{lcccccccccccccccc}
        \toprule
        Gene & \multicolumn{2}{c}{TP53 (38$\%$)} & \multicolumn{2}{c}{PIK3CA (22$\%$)} & \multicolumn{2}{c}{PTEN (14$\%$)} & \multicolumn{2}{c}{KRAS (11$\%$)} & \multicolumn{2}{c}{ARID1A (9$\%$)} & \multicolumn{2}{c}{EGFR (5$\%$)} & \multicolumn{2}{c}{FGFR2 (3$\%$)} & \multicolumn{2}{c}{Overall}\\        
        \cmidrule(lr){2-3} \cmidrule(lr){4-5} \cmidrule{6-7} \cmidrule(lr){8-9} \cmidrule(lr){10-11} \cmidrule(lr){12-13} \cmidrule(lr){14-15} \cmidrule(lr){16-17} 
        Backbone\texttt{\textbackslash}Metric & {AUC($\%$)} & {F1($\%$)}  & {AUC($\%$)} & {F1($\%$)}  & {AUC($\%$)} & {F1($\%$)}  & {AUC($\%$)} & {F1($\%$)}  & {AUC($\%$)} & {F1($\%$)} & {AUC($\%$)} & {F1($\%$)}  & {AUC($\%$)} & {F1($\%$)} & {OAUC($\%$)} & {OF1($\%$)}\\ \midrule

        GeneHe-VE\cite{qu2021genetic} & 71.6\tiny{$\pm6.2$} & 63.1\tiny{$\pm$4.6}  & 57.2\tiny{$\pm$8.3} & 50.7\tiny{$\pm$5.1}  & 66.6\tiny{$\pm$9.7} & 58.7\tiny{$\pm$5.7}  & 65.2\tiny{$\pm$9.2} & 50.4\tiny{$\pm$6.1}  &  57.7\tiny{$\pm$12.7} & 44.2\tiny{$\pm$7.1} & 57.0\tiny{$\pm$13.9} & 33.9\tiny{$\pm$9.3}  & 55.3\tiny{$\pm$14.1} & 30.1\tiny{$\pm$11.7}  & 55.2\tiny{$\pm$8.1} & 31.2\tiny{$\pm$0.8}\\
        DeepHis-VE\cite{kather2020pan} & 72.2\tiny{$\pm$6.4} & 63.9\tiny{$\pm$4.8}  & 65.8\tiny{$\pm$9.3} & 56.2\tiny{$\pm$5.9}  & 72.5\tiny{$\pm$10.7} & 61.5\tiny{$\pm$6.5} & 77.6\tiny{$\pm$11.3} & 69.1\tiny{$\pm$7.4}  & 69.8\tiny{$\pm$9.8} & 56.8\tiny{$\pm$8.7}  & 70.0\tiny{$\pm$18.1} & 49.1\tiny{$\pm$10.7}  & 61.2\tiny{$\pm$18.3} & 41.4\tiny{$\pm$12.3}  & 55.4\tiny{$\pm$8.8} & 31.9\tiny{$\pm$0.8} \\
        Att-VE\cite{ilse2018attention} & 72.3\tiny{$\pm$5.9} & 64.4\tiny{$\pm$3.6}  & 63.7\tiny{$\pm$6.3} & 53.0\tiny{$\pm$5.7}  &62.8\tiny{$\pm$7.7} & 52.7\tiny{$\pm$6.1}  & 72.5\tiny{$\pm$10.3} & 51.1\tiny{$\pm$6.7}  & 52.0\tiny{$\pm$13.9} & 43.7\tiny{$\pm$7.7}  & 73.3\tiny{$\pm$12.3} & 56.1\tiny{$\pm$9.0}  & 66.9\tiny{$\pm$14.5} & 45.3\tiny{$\pm$11.0}  & 61.2\tiny{$\pm$4.6} & 34.3\tiny{$\pm$0.6}  \\
        Trans-VE\cite{shao2021transmil} & 74.5\tiny{$\pm$5.7} & 69.6\tiny{$\pm$3.6} & 62.9\tiny{$\pm$6.6} & 55.3\tiny{$\pm$4.3}  & 68.3\tiny{$\pm$7.1} & 56.5\tiny{$\pm$5.1} & 74.5\tiny{$\pm$8.0} & 51.5\tiny{$\pm$6.1}  & 66.9\tiny{$\pm$9.1} & 52.7\tiny{$\pm$7.3}  & 60.4\tiny{$\pm$11.8} & 43.4\tiny{$\pm$9.3}	& 62.6\tiny{$\pm$12.9} & 41.9\tiny{$\pm$10.9} & 63.8\tiny{$\pm$4.4} & 36.3\tiny{$\pm$0.5}\\
        KAT\cite{zheng2022kernel} & 72.7\tiny{$\pm$5.3} & 68.3\tiny{$\pm$4.0} & 66.1\tiny{$\pm$5.8} & 58.9\tiny{$\pm$4.9}  & 78.1\tiny{$\pm$6.3} & 73.0\tiny{$\pm$4.7}  & 77.6\tiny{$\pm$7.6} & 53.6\tiny{$\pm$6.3}  & 56.2\tiny{$\pm$12.5} & 48.3\tiny{$\pm$8.1}  & 71.9\tiny{$\pm$14.9} & 48.8\tiny{$\pm$10.5}  & 57.0\tiny{$\pm$15.8} & 36.3\tiny{$\pm$13.4}  & 60.3\tiny{$\pm$3.3} & 36.1\tiny{$\pm$0.5}\\    
        HIPT\cite{chen2022scaling} & 73.3\tiny{$\pm$6.1} & 69.1\tiny{$\pm$3.7} & 66.8\tiny{$\pm$6.4} & 60.1\tiny{$\pm$4.7}  & 71.2\tiny{$\pm$6.1} & 67.2\tiny{$\pm$5.1} & 80.9\tiny{$\pm$8.3} & 60.4\tiny{$\pm$7.0}  & 70.8\tiny{$\pm$13.9} & 63.3\tiny{$\pm$9.3}  & 62.6\tiny{$\pm$10.1} & 46.0\tiny{$\pm$10.8}  & 67.5\tiny{$\pm$11.9} & 44.2\tiny{$\pm$12.9} & 64.8\tiny{$\pm$5.2} & 36.0\tiny{$\pm$0.5} \\
        \bottomrule
    \end{tabular}}
\end{table*}
\begin{table*}[htbp]
\centering \small
\caption{Ablation study showing the effectiveness of our KAM compared with other graph aggregation modules across various visual extractor backbones. Results are shown as mean $\pm$ standard error of the overall F1 score ($\%$). $\dag$ denotes a statistically significant improvement, i.e., a $p$ value below 0.005 in the student's t-test.}\label{tab3}
    \addtolength{\tabcolsep}{0.3pt}
\begin{tabularx}{\textwidth}{lcccccc} 
    \toprule
  Method\texttt{\textbackslash}Backbones  & GeneHe-VE\cite{qu2021genetic}  & DeepHis-VE\cite{kather2020pan} & Att-VE\cite{ilse2018attention}  & Trans-VE\cite{shao2021transmil} & KAT\cite{zheng2022kernel} & HIPT\cite{chen2022scaling} \\ 
    
    \midrule
    % MLP classifier & 0.248          & 0.254          & 0.256          & 0.306          & 0.298          & 0.301          & 0.339(35.7/32.3)           \\
    w/o KAM & 25.3 $\pm$ 1.5 ($\dag$) & 24.8 $\pm$ 0.8 ($\dag$) & 25.9 $\pm$  1.0 ($\dag$) & 27.3 $\pm$ 1.2 ($\dag$) & 27.1 $\pm$ 1.2 ($\dag$) & 28.3 $\pm$ 0.9  ($\dag$)\\
    
    GCN\cite{kipf2016semi}  & 30.6$\pm$ 0.8   & 30.1$\pm$ 0.8 ($\dag$)   & 29.6 $\pm$ 0.7 ($\dag$) & 32.3 $\pm$ 0.6 ($\dag$)  & 32.9 $\pm$ 0.5 ($\dag$)   & 34.1 $\pm$ 0.5 ($\dag$) \\
    
    Graph Transformer\cite{dwivedi2020generalization} & 31.0 $\pm$ 0.9    & 31.6 $\pm$ 1.0    & 30.3 $\pm$ 0.8  ($\dag$)     & 32.7 $\pm$  0.7   ($\dag$)      & 34.7  $\pm$ 0.6  ($\dag$)      & 35.4 $\pm$ 0.6   ($\dag$)   \\
    MCAT\cite{chen2021multimodal} & 30.8 $\pm$ 0.8  & 30.7 $\pm$ 0.8 ($\dag$) & 31.6 $\pm$  0.6 ($\dag$) & 32.5 $\pm$  0.5 ($\dag$) & 33.2 $\pm$  0.5 ($\dag$) & 34.4 $\pm$  0.7  ($\dag$)\\

    \midrule
    KAM (Ours)   & \textbf{31.2} $\pm$ 0.8 & \textbf{31.9} $\pm$ 0.8  & 
    \textbf{34.3} $\pm$ 0.6 & \textbf{36.3} $\pm$ 0.5 & \textbf{36.4} $\pm$ 0.5         & \textbf{36.0} $\pm$ 0.5 \\
    
    \bottomrule
\end{tabularx}
\end{table*}
\subsection{Comparison experiment on SOTA methods}
\begin{figure*}[ht]
% \begin{figure}
    \centering
    \begin{subfigure}[b]{0.33\textwidth}
        % \centering
        \includegraphics[width=\linewidth]{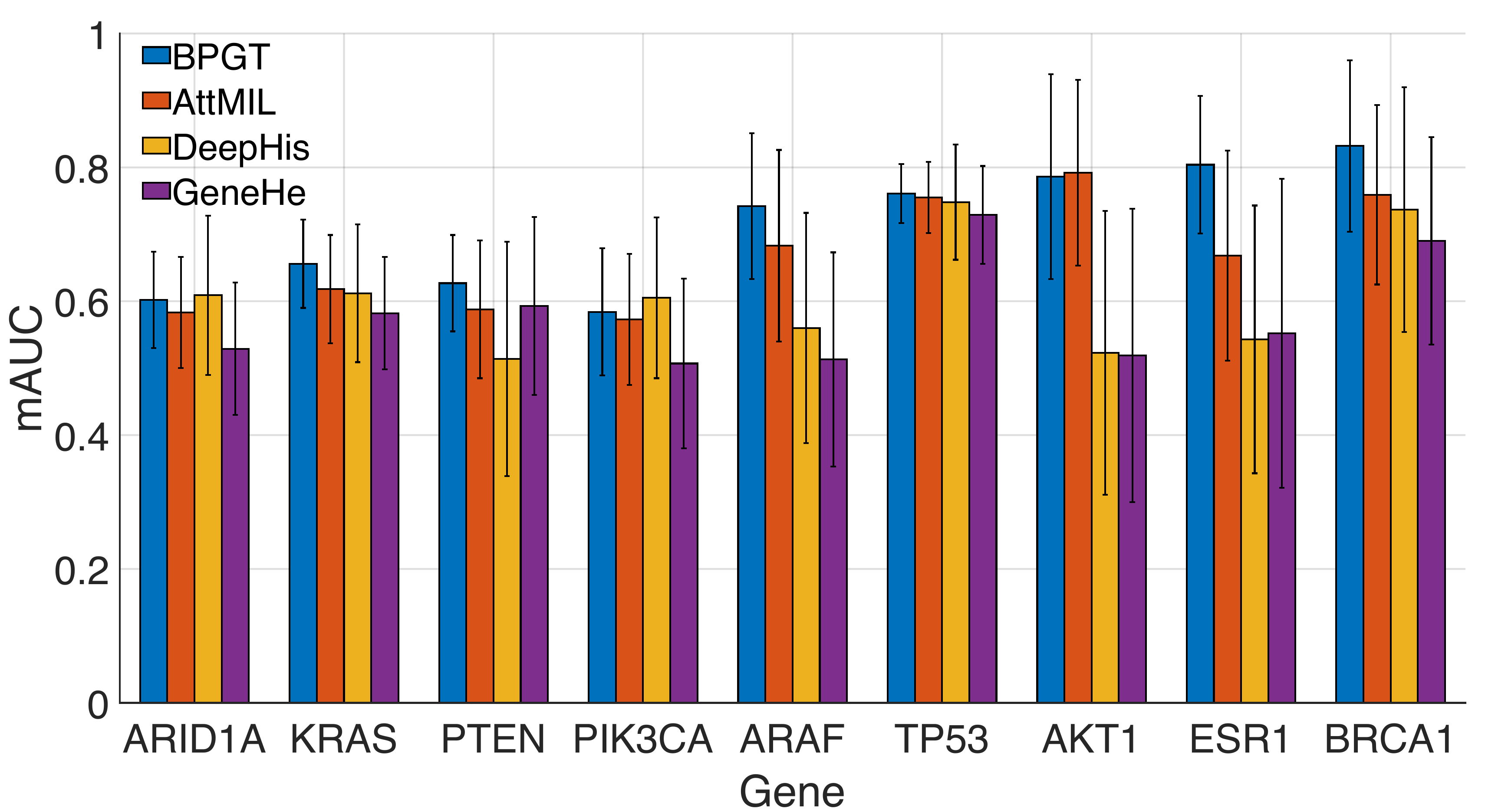}
        \caption{Performance Comparison on BRCA.}
        \label{fig_4}
    \end{subfigure}
    % \hfill
    \begin{subfigure}[b]{0.33\textwidth}
        % \centering
        \includegraphics[width=\linewidth]{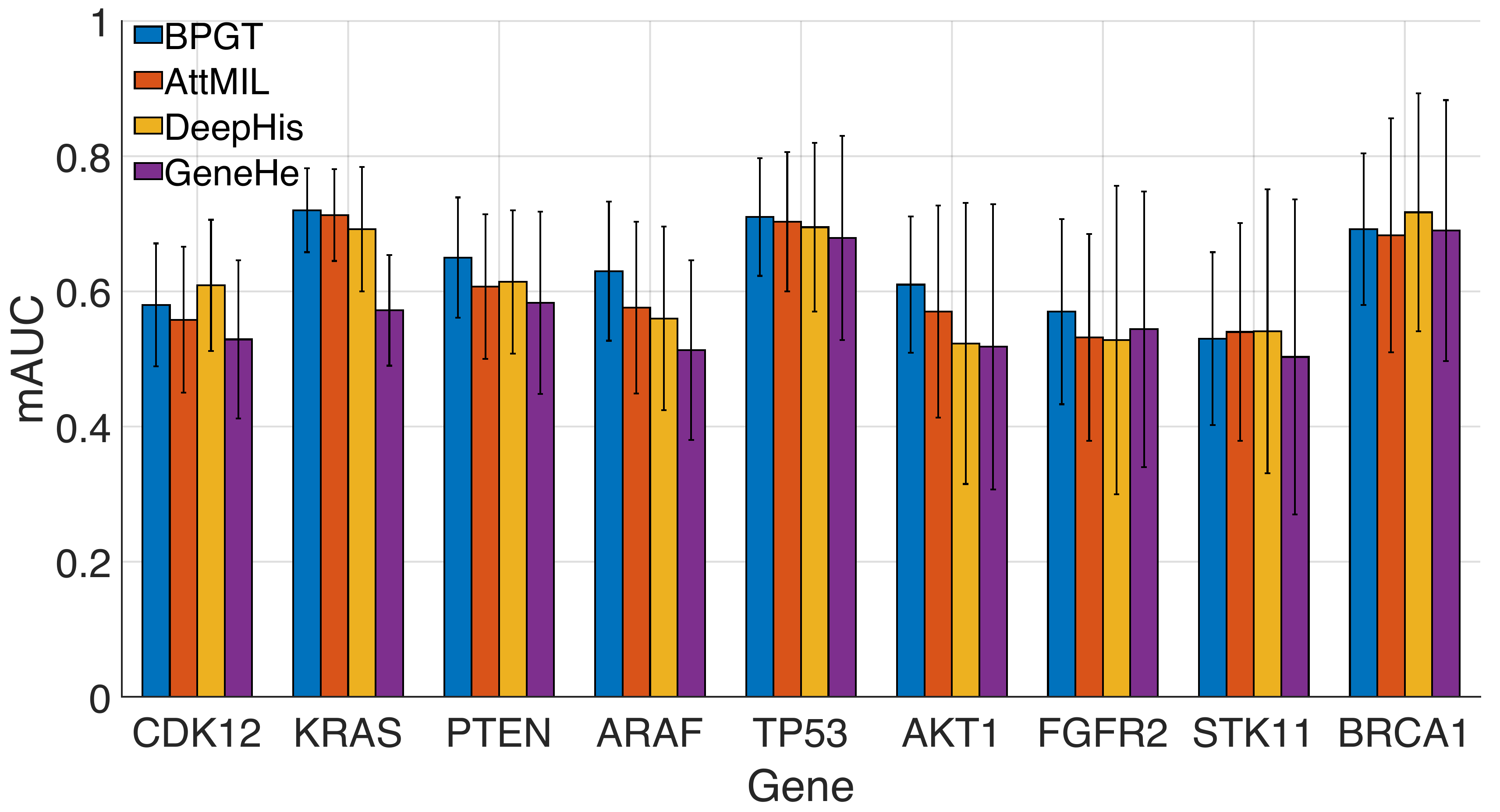}
        \caption{Performance Comparison on UCEC.}
        \label{fig_9}
    \end{subfigure}
    % \hfill    
    \begin{subfigure}[b]{0.33\textwidth}
        % \centering
        \includegraphics[width=\linewidth]{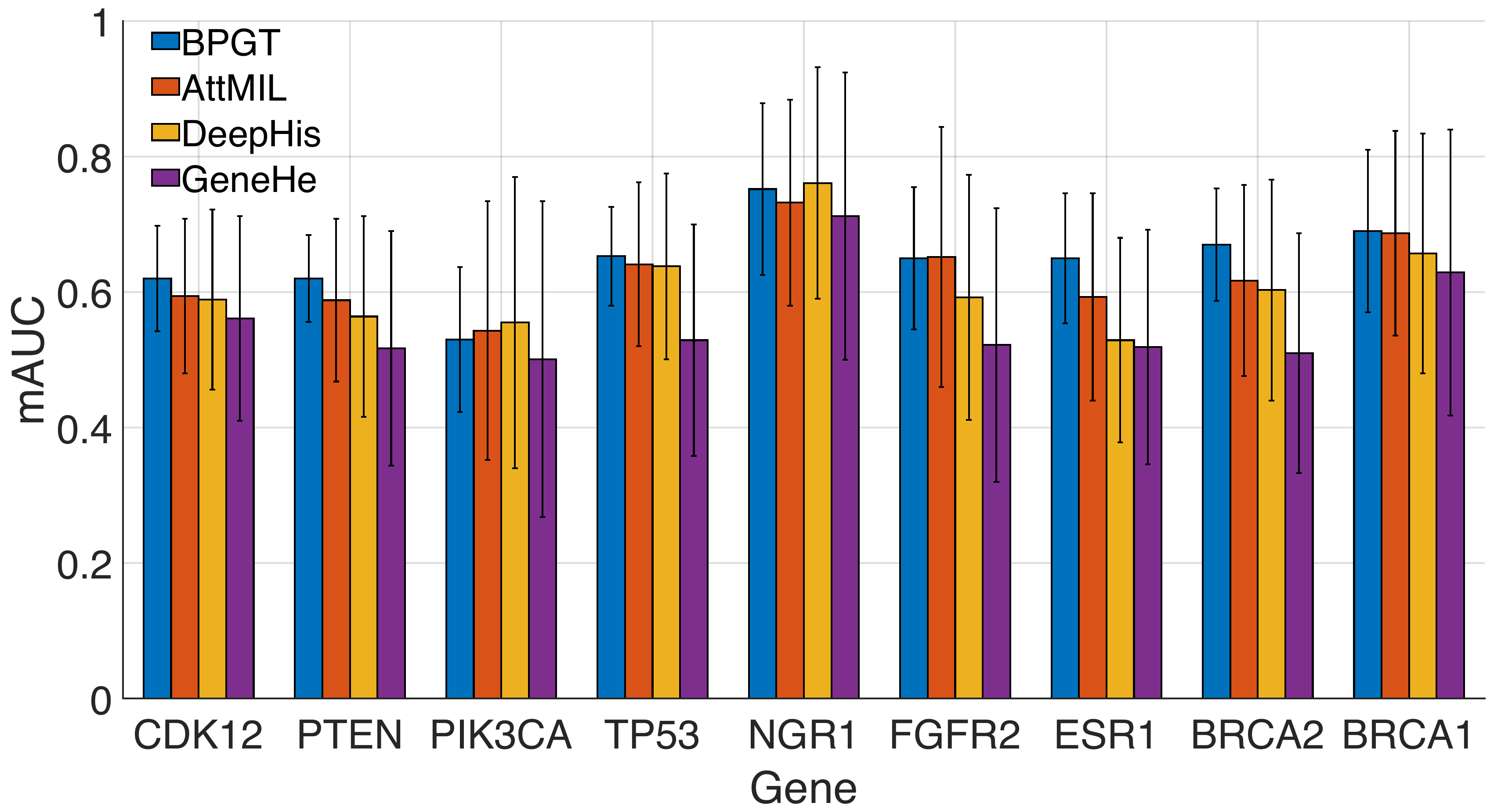}
        \caption{Performance Comparison on HNSC.}
        \label{fig_10}
    \end{subfigure}
    \caption{Performance comparison of BOGT and SOTA methods. Results are shown by the mean AUC performances on the 5-fold cross-validation for different genes on different cancers. The bars of different colors in Fig. \ref{fig:cancer} (a)$\sim$(c) represent the mean AUC of different models.}
    \label{fig:cancer}
    \vspace{-20pt}
\end{figure*}

To demonstrate the competitive performance of the BPGT model, we compare the SOTA binary-classification genetic mutation models, including GeneHe\cite{qu2021genetic}, DeepHis\cite{kather2020pan}, and AttMIL\cite{saldanha2023self}. Note that the notations ``GeneHe'' and ``DeepHis'' differ from the backbones ``GeneHe-VE'' and ``DeepHis-VE'' marked with ``-VE'' in Table. ~\ref{tab12} The former means directly using the methods in \cite{qu2021genetic}, \cite{kather2020pan}, and \cite{saldanha2023self} to train classifiers for each key gene individually, whereas the latter means employing the visual extractors from the referenced methods, which are then integrated with the GE and LD in our BPGT to perform multi-label classification for all genes. To demonstrate the advantages of our approach over SOTA methods, we compare the classification performances of key genes across various cancer types, as illustrated in Fig.~\ref{fig:cancer}. Specifically, Fig.~\ref{fig:cancer} (a), (b), and (c) detail the performance comparisons (measured by the mean AUC) for breast cancer (BRCA), uterine corpus endometrial carcinoma (UCEC), and head and neck squamous cell carcinoma (HNSC), respectively.
BPGT consistently achieves higher mean AUC values across nearly all genes listed. For example, in the case of the ARID1A gene, BPGT achieves a mean AUC of 60.2\%, outperforming GeneHe's 52.9\%, and similarly leads in the case of the AKT1 gene with a score of 78.6\% compared to GeneHe's 51.9\%. This trend continues with the other genes. Furthermore, it is shown that BPGT's standard deviations in mAUC are consistently lower than those of existing competitors, which shows that the performance of BPGT is more stable than the compared methods. For instance, in the CDK12 classification, BPGT has a standard deviation of 0.078, which is substantially lower than AttMIL's 0.114, DeepHis's 0.133, and GeneHe's 0.151. This pattern of BPGT's leading performance is replicated across other critical genes like TP53, NGR1, FGFR2, ESR1, BRCA2, and BRCA1, highlighting its robust predictive capabilities and its potential to revolutionize MIL-based gene mutation prediction methods.

\noindent\textbf{Discussion.} Our BPGT surpasses the compared approaches due to its innovative integration of biomedical and linguistic knowledge within its gene encoder. This integration is likely to produce more robust and clinically relevant predictions by leveraging the rich context provided by the combination of these knowledge domains. Furthermore, the knowledge association module within the gene encoder uses transformer-based graph representation learning to capture intrinsic relationships between mutations, which likely provides a more nuanced understanding compared to methods that do not use such a comprehensive approach. Lastly, the design of a comparative multi-label loss function not only enables the model to better differentiate between mutated and non-mutated genes but also inherently avoids the class imbalance problem in the compared methods that perform binary classification for each gene. These are the benefits of our designed BPGT framework compared with similar existing work.

\subsection{Ablation study}
\subsubsection{Impact of gene encoder}
\begin{figure}[t]
	\centering
	\includegraphics[width=\linewidth]{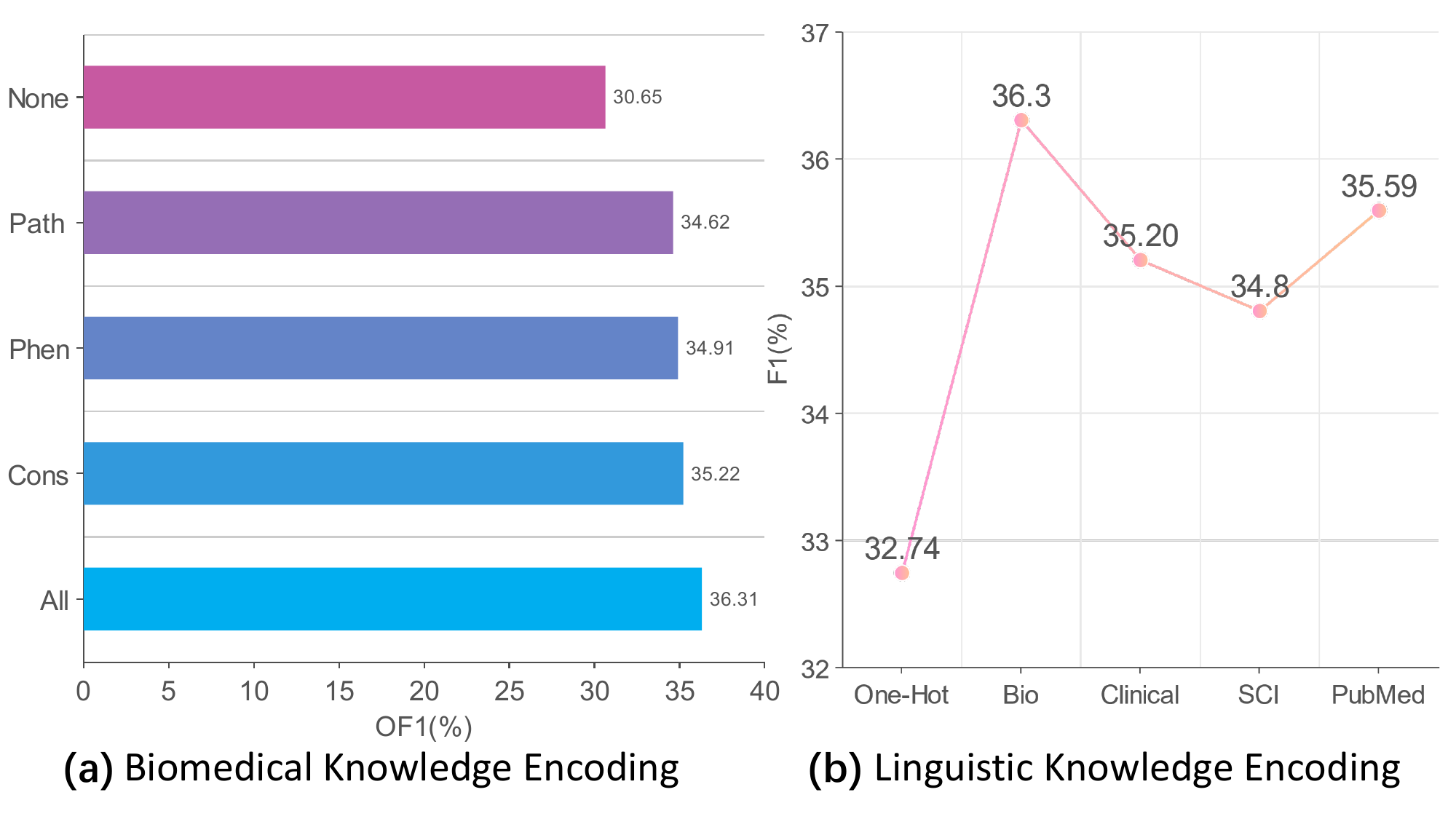}
	% \vspace{-0.8cm}
	\caption{Comparison of varying knowledge encoding approaches. a) Performance under varying biomedical knowledge encodings, i.e., without any encoding, pathway encoding, phenotype encoding, and consistency encoding. b) Performance under varying linguistic encoding approaches, i.e., one-hot encoding (which means the initial node features are one-hot labels without considering the linguistic knowledge), Bio-BERT\cite{lee2020biobert}, Clinical-BERT\cite{alsentzer2019publicly}, SCI-BERT\cite{beltagy2019scibert}, and PubMed-BERT\cite{peng2019transfer}. Results are shown as the overall F1 score.}
	\label{fig_7}
     \vspace{-10pt}
\end{figure}

\noindent\textbf{Effectiveness of linguistic and biomedical knowledge.}\label{exp.encode}
First, we perform ablation experiments to evaluate the linguistic and biomedical knowledge encoding in the GG. Trans-VE is used as the visual extractor in these experiments. As shown in Fig. \ref{fig_7} (a), the F1 scores indicate that BPGT combining all three types of biomedical knowledge (the ``All'' bar) yields the best performance with an F1 score of 36.31\%. This demonstrates that jointly incorporating biomedical knowledge (i.e., using the three encodings) benefits gene mutation classification. Then in Fig. \ref{fig_7} (b), different BERT-based models are evaluated to demonstrate the significance of linguistic knowledge. Here, Bio-BERT leads with the highest F1 score of 36.3\%, suggesting that it is the most effective linguistic encoding strategy among those tested. These results demonstrate that no matter what text encoding methods are employed to acquire linguistic knowledge, the performance consistently surpasses the ablation that does not utilize linguistic knowledge. 

\begin{figure}[t]
	\centering
	\includegraphics[width=\linewidth]{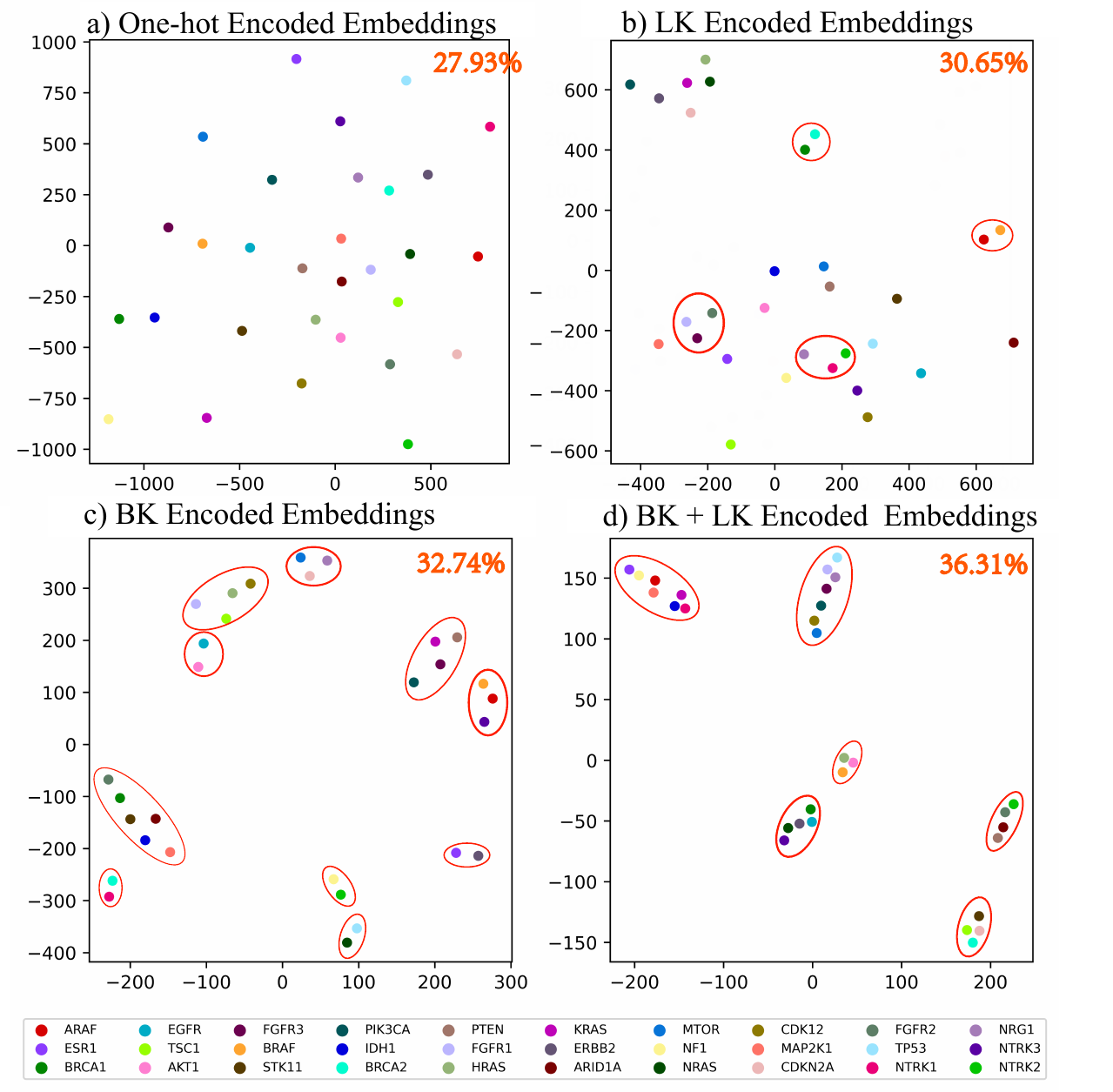}
	\caption{Visualization of the benefits brought about by the linguistic and biomedical knowledge in gene priors. The gene priors containing different knowledge are projected to 2-dimensional space by t-distributed stochastic neighbor embedding (tSNE). Fig. \ref{fig_6} (a)$\sim$(d) respectively shows the tSNE embeddings using no knowledge (i.e., one-hot embeddings), linguistic knowledge (LK), biomedical knowledge (BK), and both types of knowledge. The value on the top right of each sub-figure is the OF1 score obtained from the model that utilizes the corresponding type of knowledge. It is shown that the OF1 is higher and that genes with similar functions (e.g., ATK1, BRAF, and HRAS) are gathering together with more knowledge injected into the gene priors.}
	\label{fig_6}
	% \vspace{-0.5cm}
   \vspace{-10pt}
\end{figure}

\begin{figure*}[t]
	\centering
	\includegraphics[width=0.8\textwidth]{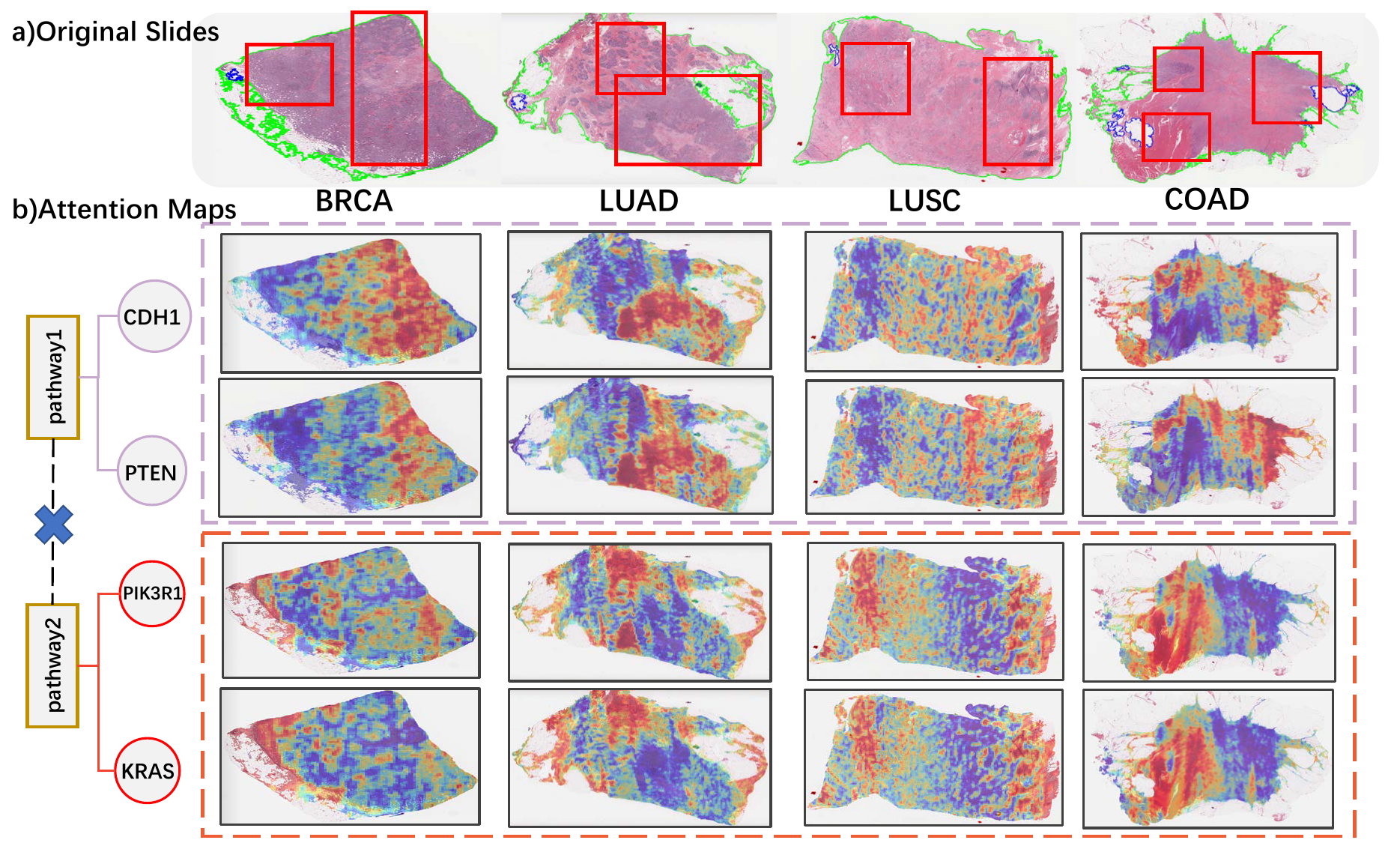}
        \caption{Visualization of attention map shows that genes that are functionally associated would highlight the same regions in the gigapixel-level WSIs. Fig. \ref{fig_5} (a) shows the original slides of patients with different cancer types; the red boxes mark the tumorous regions of each slide. Fig. \ref{fig_5} (b) shows four genes from two pathways (CDHl and PTEN belong to pathway 1, whereas PIK3CA and KRAS belong to pathway 2). Genes from the same pathways (i.e., functionally associated genes) highlight the same WSI regions that are highly aligned with the tumorous regions in Fig. \ref{fig_5} (a), which gives evidence that LD leverages gene priors to focus on genetic mutation-related regions in the WSIs. }	
	\label{fig_5}
	\vspace{-0.2cm}
\end{figure*}
\noindent\textbf{Effectiveness of knowledge association module.}
To verify the effectiveness of the KAM, we carry out ablation experiments with no knowledge association (i.e., using only the initial gene features $\mathbf{G}$ as gene priors, as shown in the first row of Table. ~\ref{tab3}) and experiments with different graph-based feature aggregation modules, including GCN\cite{kipf2016semi}, graph transformer\cite{dwivedi2020generalization}, and MCAT \cite{chen2021multimodal} (as shown in the second to fourth rows of Table.~\ref{tab3}). 
The comparisons in Table.~\ref{tab3} indicate that associating knowledge from different modalities (i.e., linguistic knowledge and biomedical knowledge) benefits genetic mutation prediction, with KAM exceeding the performance of all other knowledge association methods across all visual extractor backbones. 
For example, the first row without knowledge association results in the lowest performance, which verifies the importance of knowledge association. Moreover, KAM achieves a significant performance boost with an F1 score of 36.3\% ± 0.5 with the Trans-VE backbone, compared to the next best, graph transformer, which scores 32.7\% ± 0.7. 
Furthermore, the superiority of KAM over other graph aggregation modules is statistically significant in most comparisons (the annotation "$\dag$" denotes a p-value smaller than 0.005). In some rare cases (such as cases where the GeneHe-VE and DeepHis-VE are used), the improvements brought about by the KAM are not that significant (although, the mean OF1 values are still higher using the gene priors constructed by KAM). This could be attributed to the fact that these two visual encoders only utilize a subset of WSI patches, which may not provide sufficient visual information and limit the functions of the gene priors. As a comparison, for visual encoders that fully leverage WSI patches, the advantages of KAM become more pronounced. In all, KAM associates knowledge from multiple modalities in a transformer-based graph representation learning manner, which is demonstrated to outperform the ablated versions in gene mutation classification.

\begin{table}[ht]
    \centering
    \caption{Ablation study showing the effectiveness of the components in label decoder. The visual extractor is selected as TransVE\cite{shao2021transmil}. Results (mean $\pm$ standard deviation) are evaluated based on 5-fold cross-validation.}\label{tab10}

\addtolength{\tabcolsep}{5.5pt}
\begin{tabularx}{0.95\linewidth}{l|cc}
%	\begin{tabular}{lcccc}
		\toprule
		Architecture & Parameters & OF1($\%$) \\ 
        \midrule
        w/o modality fusion module & 2,507,258 & 28.33 $\pm$ \tiny{1.80} \\	
        w/o multi-label loss & 2,902,970 & 23.37 $\pm$ \tiny{0.59} \\	
        
		\midrule
        $\textbf{BPGT}$ & 2,902,970 & 36.38 $\pm$ \tiny{0.53} \\
		\bottomrule
%	\end{tabular}
\end{tabularx}

\end{table}
\noindent\textbf{Discussion.}
The linguistic and biomedical knowledge encodings in the gene priors provide complementary gene information, which improves the discriminability of gene features and benefits mutation prediction performance. This is also proven in Fig. \ref{fig_6}, which visualizes the gene priors containing different types of knowledge by the t-distributed stochastic neighbor embedding (tSNE) that projects them to the 2-dimensional tSNE map. Values on the top right of sub-figures are the OF1 scores obtained from models that utilize the corresponding type of knowledge. 
Fig. \ref{fig_6} (a) illustrates gene priors only employing one-hot encoding, i.e., without leveraging either linguistic knowledge (LK) or biomedical knowledge (BK) for gene representation, which results in an OF1 score of 27.93\%. Fig.~\ref{fig_6} (b) and (c) depict gene priors utilizing solely LK and BK, respectively achieving OF1 scores of 30.65\% and 32.74\%. Fig.~\ref{fig_6} (d) demonstrates the gene priors that integrate both LK and BK, which achieves the highest OF1 score of 36.31\%. 
By comparing with Fig. \ref{fig_6} (a) and (b), the introduction of LK may capture gene functions, interactions, and regulatory mechanisms, which prompts the linguistically associated genes to be positioned closer while unrelated genes remain separated. For instance, BRCA1 and BRCA2 genes are clustered together since they share similar nomenclature “BRC”; ARAF and BRAF genes are clustered together since they share similar nomenclature “A-Raf Proto-Oncogene” and “B-Raf Proto-Oncogene”, respectively.
Similarly, by comparing Fig. \ref{fig_6} (a) and (c), the introduction of BK brings in empirical data from biomedical research such as gene-disease associations and pathway participation to boost the prediction OF1. For instance, the mutation of ESR1 and ERBB2 are both closely associated with breast cancer and are both involved in multiple pathways such as the “PI3K/AKT Signaling in Cancer” pathway, “RNA Polymerase II Transcription” pathway and “Signaling by Receptor Tyrosine Kinases” pathway.
Furthermore, Fig. \ref{fig_6} (d) exhibits the highest OF1 score of 36.31\% and a more notable clustering of embeddings by the synergistic incorporation of BK and LK, compared with Fig. \ref{fig_6} (b) and (c). For instance, the AKT1 gene, BRAF gene, and HRAS gene are linguistically similar since they encode protein kinase B, protein kinase B-Raf, and protein GTPase HRas in their gene summary, respectively. These genes are also functionally related since they are all involved in the “Diseases of signal transduction by growth factor receptors and second messengers” pathway, “Oncogenic MAPK signaling” pathway, and “Signaling by high-kinase activity BRAF mutants” pathway. This indicates combining LK and BK could probably offer a more comprehensive representation of gene characteristics; they make functionally related genes cluster together and thus enhance the discrimination of gene features for downstream tasks, which benefits genetic mutation prediction and consequently improves the OF1 score.

\subsubsection{Impact of label decoder}
For LD, we separately perform ablation experiments assessing the designed MFM and the comparative multi-label loss. Results are demonstrated in Table.~\ref{tab10}. 

\noindent\textbf{Effectiveness of modality fusion module.} This ablation simply fuses the features by matrix operations in place of the modality fusion module and then adjusts the output dimension to K*1 to obtain the logits. 
As shown in Table.~\ref{tab10}, the MFM dramatically increases the OF1 score (from $28.33\%$ to $36.38\%$) and reduces the standard deviation (from $1.80\%$ to $0.53\%$). Furthermore, while the MFM leads to an increment of $\sim 8\%$ in OF1, it does not significantly increase the total number of parameters in the network. This highlights the advantage of aggregating visual and textual information in a transformer framework for multi-label classification. As will be analyzed, the MFM could assist in capturing the critical regions in the attention map, which employs the most informative visual features to predict the mutation information.

\noindent\textbf{Effectiveness of comparative multi-label loss.} This ablation aims to validate the effectiveness of comparative multi-label loss, which replaces the comparative multi-label loss with the binary-cross-entropy (BCE) loss to optimize the multi-label classification. For a fair comparison, we keep the same learning rate and training strategy. As Table.~\ref{tab10} shows, the comparative multi-label loss drastically enhances the OF1 score from $23.37\%$ to $36.38\%$ and reduces the standard deviation from $0.59\%$ to $0.53\%$. The comparative multi-label loss maintains the same model capacity yet leads to a significant enhancement in performance by the margin of OF1 of $13.01\%$, which proves the critical role of comparative multi-label loss in achieving superior performance outcomes. While the BCE loss independently compares the positive and negative logits for each class (genes), the comparative multi-label loss takes advantage of the inherent comparisons among classes (genes) within the softmax function. This key distinction naturally enables the comparative multi-label loss to alleviate the class imbalance issue that inevitably arises when using the BCE loss. As a result, the comparative multi-label loss enhances the discrimination capabilities of the BPGT for positive and negative classes, resulting in improved performance.

\noindent\textbf{Visualization of attention map for gigapixel-level WSIs} 
We visualize the attention maps from the final cross-attention layer of the LD, which depict how the model leverages gene priors to focus on critical visual features from gigapixel-level WSIs. This visualization illustrates that LD could comprehensively utilize the multi-modal information extracted from the GE to guide the model to focus on the genetic mutation-related region in gigapixel-level WSIs. 
Interestingly, the visualization demonstrates gene priors within the LD successfully capture distinct morphological features. For instance, Fig. \ref{fig_5} (a) shows the tumorous regions (shown by the red boxes) related to different cancers (annotated below the WSIs), while the high-attention regions in Fig.8(b) highly align with these regions although the red boxes are actually agnostic to BPGT. This demonstrates that the model has automatically focused on the tumorous tissues in histopathology, i.e., the LD helps BPGT to be aware of the prospective relationship between genetic mutation and tumorous regions.
Additionally, when examining the attention maps of functionally associated genes (i.e., genes belonging to the same pathways), we find that they focus on similar visual regions. In contrast, genes that are not functionally associated (i.e., genes belonging to different pathways) focus on different visual regions. For instance, as shown in the first column of Fig. \ref{fig_5} (b), CDH1 and PTEN (belonging to pathway 1) pay attention to the same WSI regions, which shows visible differences to the regions highlighted by PIK3CA and KRAS (belonging to pathway 2). The notable difference in the high-attention regions between CDH1 and PTEN (pathway 1) and PIK3CA and KRAS (pathway 2) demonstrates that the LD has leveraged gene priors such as the functional associations to find the highly correlated WSI features for genetic mutation.

\noindent\textbf{Discussion.}
As shown above, BPGT with all two designs exhibits superior performance while avoiding excessive parameter increment. The modality fusion module only introduces approximately 14\% more parameters, whereas the comparative multi-label loss has no impact on the overall parameter count. However, they lead to significant performance increments of $8.05\%$ and $13.01\%$ respectively. These results underscore the effectiveness of the modality fusion module and the comparative multi-label loss, highlighting their advanced capability in integrating biological priors with visual features from WSI and improving the discrimination of mutated and non-mutated genes.
Furthermore, the fusion of visual features and gene priors in LD prompts the attention maps' alignment with the tumorous regions. The attention maps show the model focuses on specific areas that correlate with the genetic mutations, thereby connecting the visual features with the underlying genetic information. These observations suggest that BPGT is a potential method to identify the underlying relationships between genetic mutation and histopathology features.

\section{Conclusion}
% \vspace{-0.2cm}
To the best of our knowledge, our BPGT is the first to devise the multi-label classification paradigm to predict genetic mutation. BPGT designs a gene encoder through a novel transformer-based graph representation learning approach, discovering that integrating biomedical and linguistic knowledge in the gene label helps to explore the relationships of the mutations between genes. BPGT also designs a label decoder through a transformer-based modality fusion model and a comparative multi-label loss, revealing that the underlying relationships between gigapixel-level WSIs and genetic information benefit genetic mutation prediction and that introducing comparisons among classes could better discriminate the mutated genes from the non-mutated ones in the multi-label classification paradigm.
Experiments demonstrate that the designs in BPGT can comprehensively enhance the performance of genetic mutation predictions and outperform the SOTA models. Our work could be an important step towards fully leveraging the intrinsic knowledge of genomics to improve the prediction performance of genetic mutation on patients' histopathology images. The code implementation is available at: \href{https://github.com/gexinh/BPGT.git}{https://github.com/gexinh/BPGT.git}.

%{\small
%\bibliographystyle{IEEEtran}
%% \bibliographystyle{plain}
%\bibliography{egbib}

\vspace{-40pt}
\begin{IEEEbiography}
[{\includegraphics[width=0.85in,height=1in, clip,keepaspectratio]{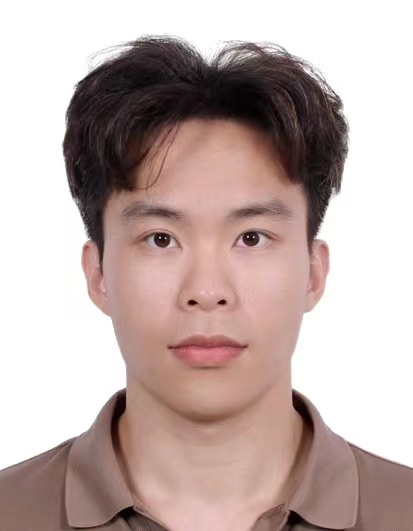}}]{Gexin Huang} is a doctoral student at the Department of Electrical and Computer Engineering, University of British Columbia. He was a research assistant at Sun Yat-Sen University. He received M.S. degree in pattern recognition and intelligent systems from the South China University of Technology in 2021. His research interests include medical imaging analysis, graph representation learning, and Bayesian deep learning.
\end{IEEEbiography}

\begin{IEEEbiography}
[{\includegraphics[width=0.85in,height=1in, clip,keepaspectratio]{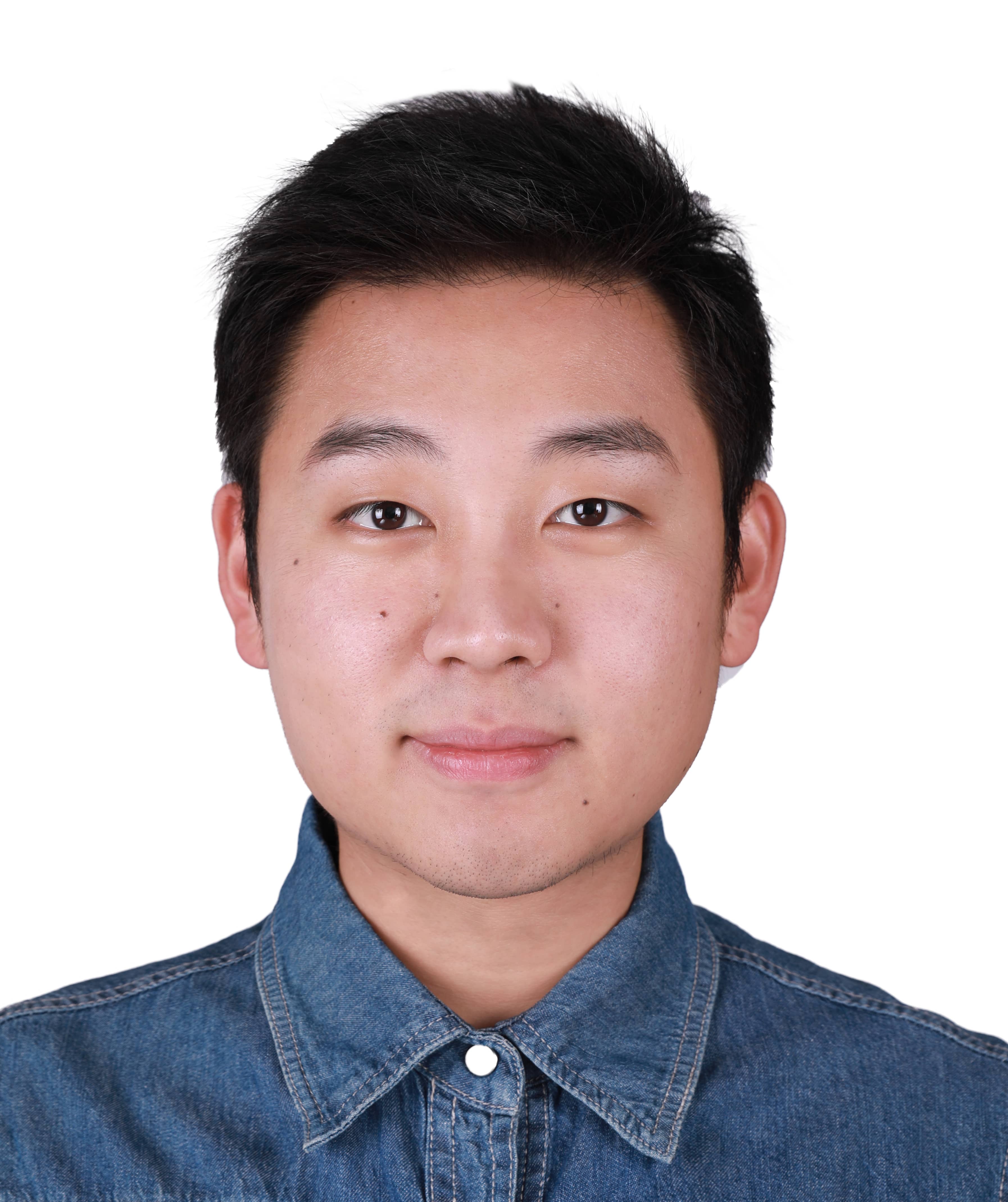}}]{Chenfei Wu} is currently a radiation oncologist in Sun Yat-sen University Cancer Center, Guangzhou, China. He received his M.D. degree from Sun Yat-sen University Cancer Center. He specializes in the diagnosis and treatment of head and neck cancer and thoracic cancer. His research primarily focuses on the application of medical AI and machine learning in radiotherapy and oncology.
\end{IEEEbiography}
\vspace{-40pt}
\begin{IEEEbiography}
[{\includegraphics[width=0.85in,height=1in, clip,keepaspectratio]{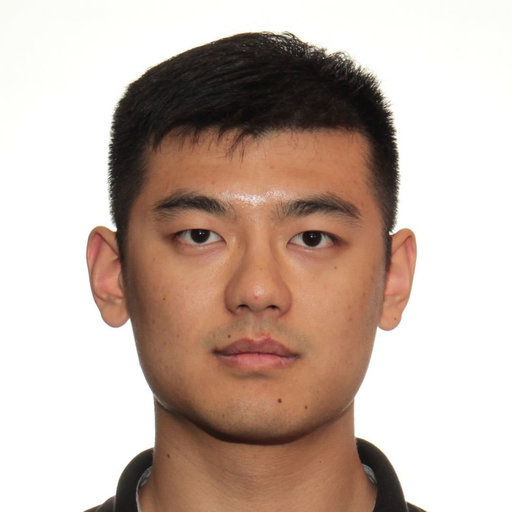}}]{Mingjie Li} is currently a post-doc research fellow at the Radiation Oncology, Stanford University. He obtained his Ph.D. degree from the University of Technology Sydney in 2023. His research interests include machine learning, medical AI, and computer vision, especially on the topic of medical vision and language alignment.
\end{IEEEbiography}
\vspace{-40pt}
\begin{IEEEbiography}
[{\includegraphics[width=0.85in,height=0.9in, clip,keepaspectratio]{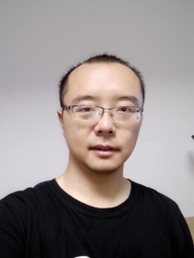}}]
{Shen Zhao} received his Ph.D. degree in Tsinghua University in 2015. He is now researching on deep learning methods in medical image analysis. His research includes novel methods on object detection/segmentation, metric learning, deep active contours, and the combination of visual and linguistic information in medical image diagnosis.
\end{IEEEbiography}
\vspace{-40pt}
\begin{IEEEbiography}[{\includegraphics[width=0.85in,height=1in,clip,keepaspectratio]{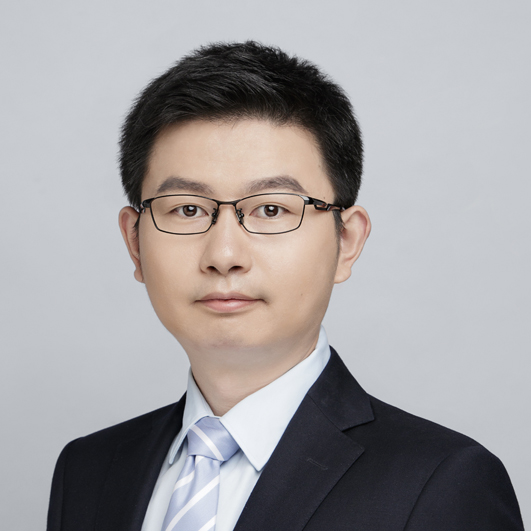}}]{Xiaojun Chang} (Senior Member, IEEE)
is a Professor at the Australian Artificial Intelligence Institute, University of Technology Sydney. Before joining UTS, he was an Associate Professor at the School of Computing Technologies, RMIT University, Australia. 
% After graduation, he subsequently worked as a Postdoc Research Fellow at School of Computer Science, Carnegie Mellon University, Lecturer and Senior Lecturer in the Faculty of Information Technology, Monash University, Australia. 
He has spent most of his time working on exploring multiple signals for automatic content analysis in unconstrained or surveillance videos. He has achieved top performances in various international competitions.
\end{IEEEbiography}
\vspace{-40pt}
\begin{IEEEbiography}[{\includegraphics[width=0.85in,height=1in,clip,keepaspectratio]{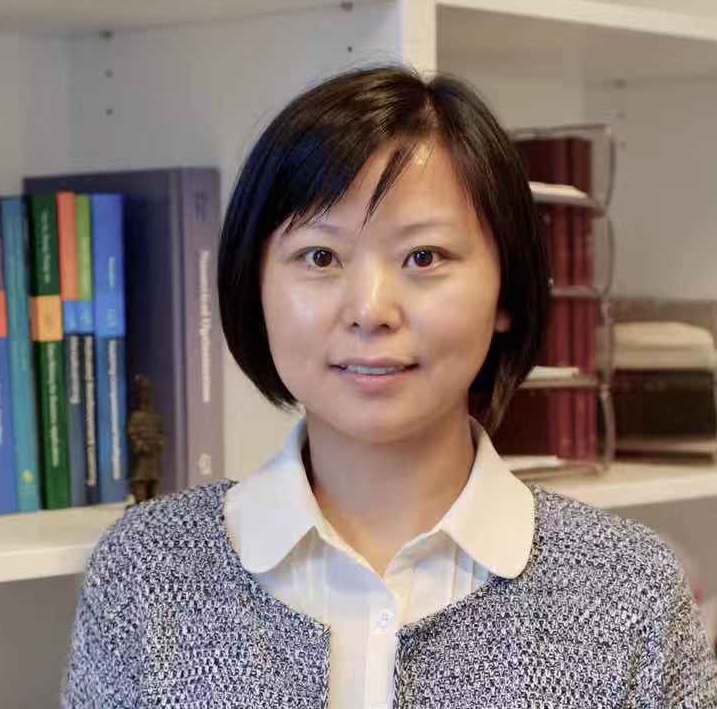}}] {Ling Chen} is a Professor with the Australian Artificial Intelligence Institute, University of Technology Sydney. She received PhD from Nanyang Technological University, Singapore. Her research area is machine learning and data mining. Her recent research focuses on anomaly detection from complex structured data, hashing and representation learning for various types of data, and reinforcement learning in text-based interactive systems. 
% Her papers appear in major journals and conferences including IEEE TPAMI, IEEE TNNLS, NeurIPS and IJCAI. % She is an editorial member of the Elsevier Journal of Data and Knowledge Engineering, the Springer Journal of Data Science and Analytics, and the IEEE Journal of Social Computing.
\end{IEEEbiography}
\vspace{-40pt}
\begin{IEEEbiography}[{\includegraphics[width=0.85in,height=1in,clip,keepaspectratio]{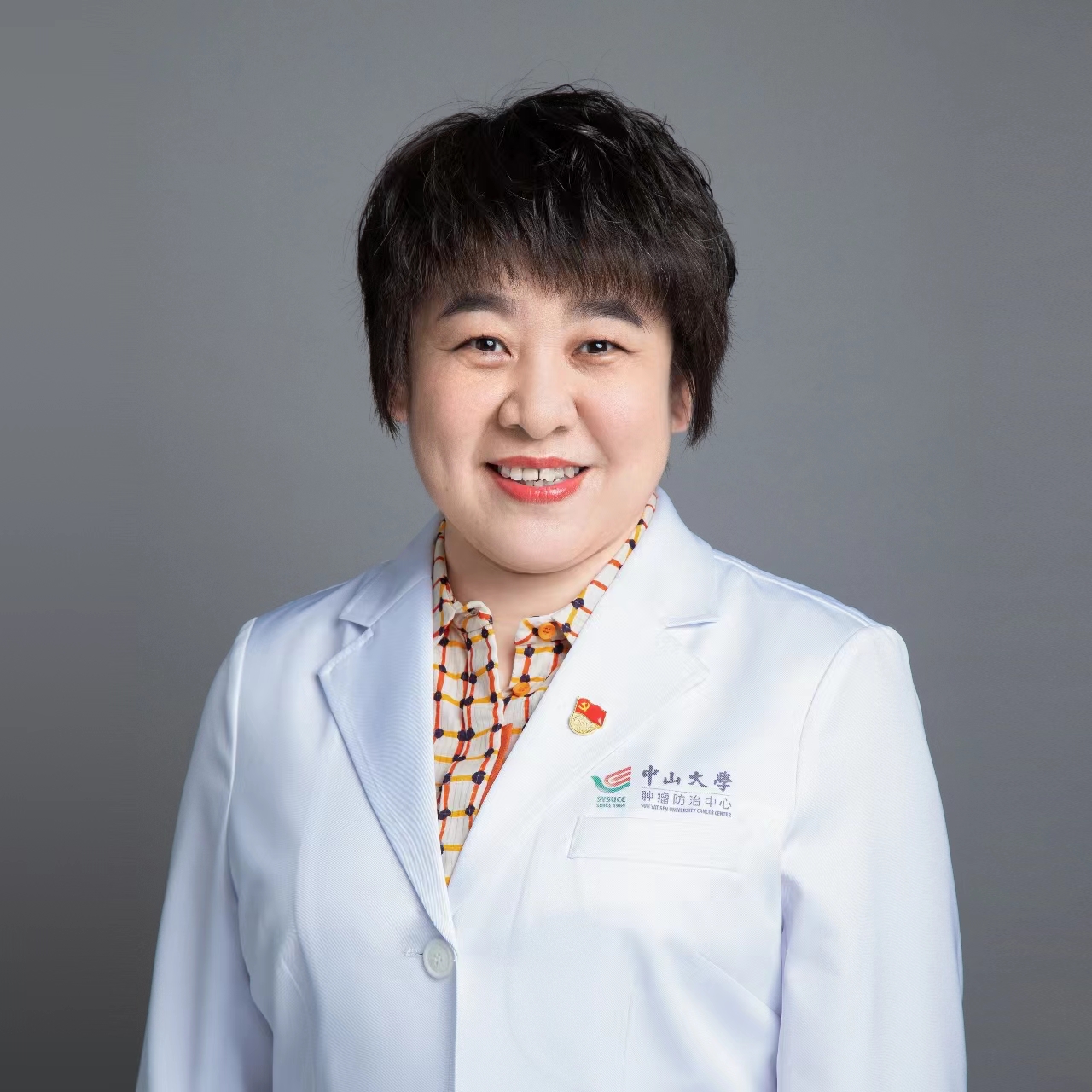}}]{Ying Sun} is a Professor of Radiation Oncology and vice president at Sun Yat-sen University Cancer Center, Guangzhou, China. Her main research interest is personalized and precision treatment of nasopharyngeal carcinoma, particularly interests include AI-assisted radiotherapy, big data-driven risk stratification, and translational research focused on developing prognostic and predictive markers. 
% She has published over 60 papers in prestigious journals, such as N Engl J Med, Lancet, Nat Med, Lancet Oncol, Nat Commun and Radiology. She is the vice Chairman of China Anti-Cancer Association-Big data and Real World Research Committee and the vice Chairman of the Chinese Society of Clinical Oncology-Nasopharyngeal Cancer Special Committee.
\end{IEEEbiography}
\vspace{-40pt}
\begin{IEEEbiography}[{\includegraphics[width=0.85in,height=1in,clip,keepaspectratio]{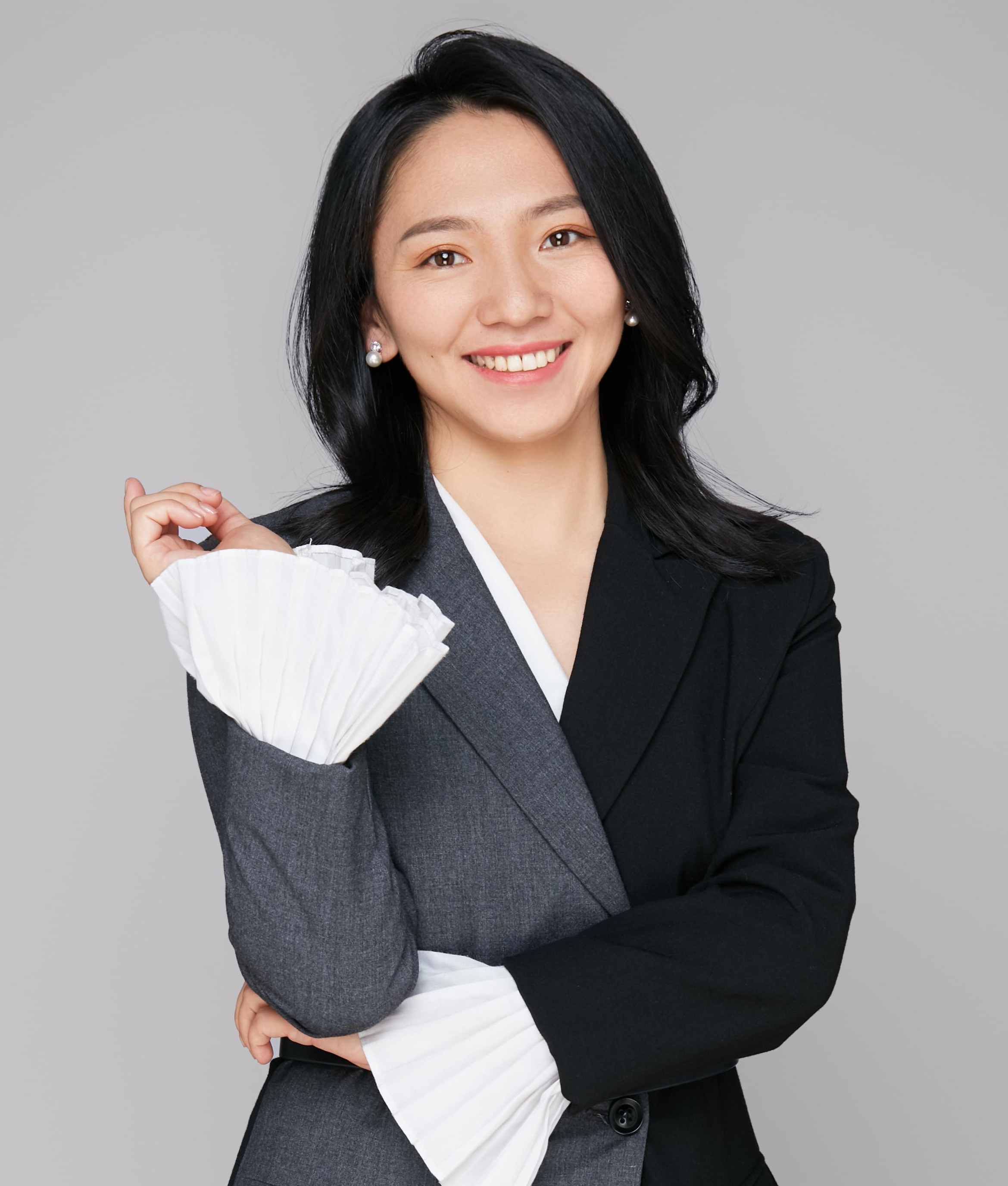}}]
{Xiaodan Liang} is currently an Associate Professor at Sun Yat-sen University. She was a Project Scientist at Carnegie Mellon University, working with Prof. Eric Xing. She received her PhD degree from Sun Yat-sen University in 2016. She has published over 80 cutting-edge papers on graph neural networks, deep reasoning, structure prediction, object detection, which have appeared in the most prestigious journals and conferences in the field, Google Citation 6000+. She serves as an Area Chair of ICCV 2019, CVPR 2020 and Tutorial Chair of CVPR 2021. She hosts the “Towards Causal, Explainable and Universal Medical Visual Diagnosis Workshop” on CVPR 2019. She also hosted the tutorial about “Structured Deep Learning for Pixel-wise Understanding” on ACM MM 2018. She has been awarded ACM Chine and CCF Best Doctoral Dissertation Award and Alibaba DAMO Academy Young Fellow. 
% She and her collaborators has also published the largest human parsing dataset to advance the research on human understanding and successfully organized the 1st Look Into Person (LIP) and 2nd Look Into Person (LIP) workshop and challenge on CVPR 2017 and CVPR 2018. 
% She served as the guest editors of Pattern Recognition Letters and journal of Multimedia Tools and Applications. She obtained 2018 ACM China Best Doctoral Dissertation Award and CCF Best Doctoral Dissertation Award. She won the 2nd place in Key Points Detection of Apparel Track of FashionAI Global Challenge 2018-Alibaba Cloud.

\end{IEEEbiography}

\vspace{-40pt}
\begin{IEEEbiography}[{\includegraphics[width=0.85in,height=1in,clip,keepaspectratio]{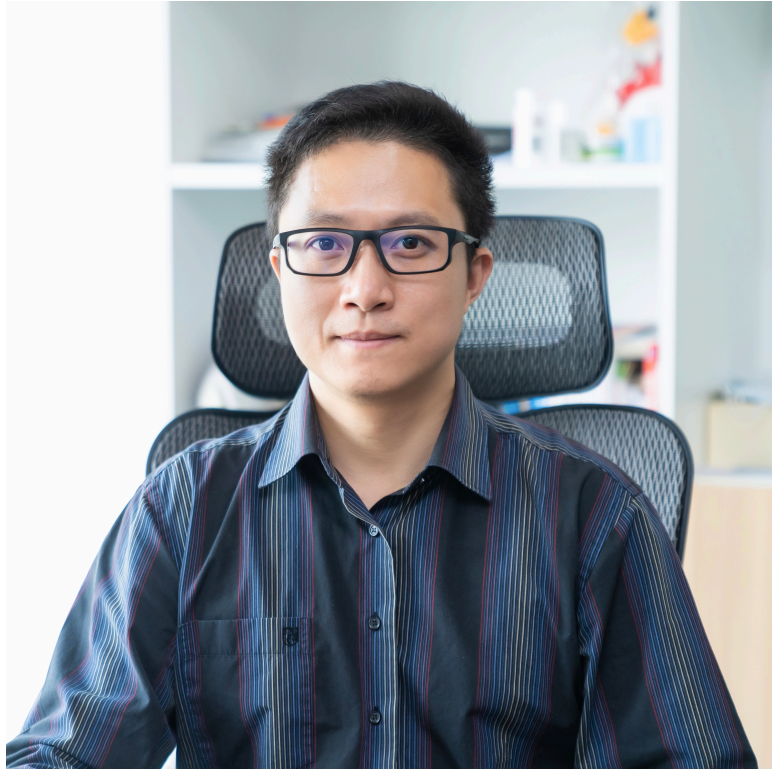}}]
{Liang Lin} is a Full Professor of Sun Yat-sen University. He served as the Executive R$\&$D Director and Distinguished Scientist of SenseTime Group from 2016 to 2018, taking charge of transferring cutting-edge technology into products. He has authored or co-authored more than 200 papers in leading academic journals and conferences. He is an associate editor of IEEE Trans, HumanMachine Systems, and IET Computer Vision. He served as Area Chair for numerous conferences such as CVPR, ICCV, and IJCAI. He is the recipient of numerous awards and honors including Wu WenJun Artifcial Intelligence Award, ICCV Best Paper Nomination in 2019, Annual Best Paper Award by Pattern Recognition (Elsevier) in 2018, Best Paper Dimond Award in IEEE ICME 2017, and Google Faculty Award in 2012. He is a Fellow of IET.
\end{IEEEbiography}
% \clearpage
% \newpage
% \appendix
% \input{appendix}

\end{document}